\documentclass[journal]{IEEEtran}
\usepackage{amsmath,amsfonts,amssymb}
\usepackage{graphicx}
\usepackage{cite}
\usepackage{booktabs}
\usepackage{algorithm}
\usepackage{algorithmic}
\usepackage{url}
\usepackage{multirow}
\usepackage{color}

\ifCLASSOPTIONcompsoc
  \usepackage[caption=false,font=normalsize,labelfont=sf,textfont=sf]{subfig}
\else
  \usepackage[caption=false,font=footnotesize]{subfig}
\fi

\usepackage{hyperref}
\hypersetup{
    colorlinks=true,
    linkcolor=blue, 
    citecolor=blue,
    urlcolor=blue,
}

\begin{document}


\title{SKANet: A Cognitive Dual-Stream Framework with Adaptive Modality Fusion for Robust Compound GNSS Interference Classification}

\author{Zhihan Zeng, \IEEEmembership{Graduate Student Member, IEEE}, Yang Zhao, \IEEEmembership{Member, IEEE},\\
Kaihe Wang, \IEEEmembership{Graduate Student Member, IEEE}, 
Dusit Niyato, \IEEEmembership{Fellow, IEEE}, Hongyuan Shu,\\
Junchu Zhao, Yanjun Huang, Yue Xiu, \IEEEmembership{Member, IEEE}, Zhongpei Zhang, Ning Wei, \IEEEmembership{Member, IEEE}%
\thanks{Zhihan Zeng, Yue Xiu, Zhongpei Zhang and Ning Wei are with the National Key Laboratory of Wireless Communications, University of Electronic Science and Technology of China, Chengdu 611731, China (E-mail: 202511220608@std.uestc.edu.cn, xiuyue12345678@163.com, zhangzp@uestc.edu.cn, wn@uestc.edu.cn). Y. Zhao is with the Nanyang Technological University, Singapore 639798 (e-mail: zhao0466@e.ntu.edu.sg). D. Niyato is with the College of Computing and Data Science, Nanyang Technological University, Singapore 639798 (e-mail: DNIYATO@ntu.edu.sg). Kaihe Wang is with the University of Electronic Science and Technology of China, Chengdu 611731, China (E-mail: khewang@yeah.net). Hongyuan Shu and Junchu Zhao are with Shanghai Aerospace Electronic Technology Institute, Shanghai 201109, China, and also with Shanghai Key Laboratory of Collaborative Computing in Spacial Heterogeneous Networks (CCSN), Shanghai 201109, China (E-mail: shuhongyuan2020@163.com, zhaojunchuhit@163.com). Yanjun Huang is with Shanghai Xiaoyuan Innovation center, China (E-mail: poer1985@163.com).
The corresponding author is Ning Wei.}}

\maketitle

\begin{abstract}
As the electromagnetic environment becomes increasingly complex, Global Navigation Satellite Systems (GNSS) face growing threats from sophisticated jamming interference. Although Deep Learning (DL) effectively identifies basic interference, classifying compound interference remains difficult due to the superposition of diverse jamming sources. Existing single-domain approaches often suffer from performance degradation because transient burst signals and continuous global signals require conflicting feature extraction scales. We propose the Selective Kernel and Asymmetric convolution Network(SKANet), a cognitive deep learning framework built upon a dual-stream architecture that integrates Time-Frequency Images (TFIs) and Power Spectral Density (PSD). Distinct from conventional fusion methods that rely on static receptive fields, the proposed architecture incorporates a Multi-Branch Selective Kernel (SK) module combined with Asymmetric Convolution Blocks (ACBs). This mechanism enables the network to dynamically adjust its receptive fields, acting as an adaptive filter that simultaneously captures micro-scale transient features and macro-scale spectral trends within entangled compound signals. To complement this spatial-temporal adaptation, a Squeeze-and-Excitation (SE) mechanism is integrated at the fusion stage to adaptively recalibrate the contribution of heterogeneous features from each modality. Evaluations on a dataset of 405,000 samples demonstrate that SKANet achieves an overall accuracy of 96.99\%, exhibiting superior robustness for compound jamming classification, particularly under low Jamming-to-Noise Ratio (JNR) regimes.
\end{abstract}


\section{Introduction}
\label{sec:introduction}

\IEEEPARstart{F}{uture} 6G networks envision a Space-Air-Ground Integrated Network (SAGIN) to achieve global connectivity, integrating terrestrial cellular systems, aerial platforms, and satellite constellations into a unified framework\cite{ref1, ref3}. Within this architecture, Global Navigation Satellite Systems (GNSS) serve two essential functions. Beyond their traditional role in positioning and navigation for Intelligent Transportation Systems (ITS) and autonomous driving, GNSS provides the spatiotemporal backbone for infrastructure \cite{ref2}. It delivers the precise timing synchronization essential for phase alignment in 5G/6G base stations, smart power grid regulation, and Internet of Things (IoT) coordination. Thus, GNSS signal reliability is critical not just for user Quality of Service (QoS) \cite{zhao2025generative,11353414,11355857,11316665,11098592,11220909,11207524,11346858,11316633,11108293,10797657}, but for the stability of cognitive networks \cite{ref4}.

However, the satellite downlink remains vulnerable. Originating from Medium Earth Orbit (MEO) satellites approximately 20,000 km above the Earth, GNSS signals reaching the surface are extremely weak, often falling below the thermal noise floor \cite{ref6}. This renders receiver front-ends highly sensitive to Radio Frequency Interference (RFI). In increasingly congested spectrums, an operating environment has shifted from simple, unintentional noise to complex disruptions driven by crowding \cite{ref7}. The widespread use of Personal Privacy Protection Devices (PPDs) and programmable Software-Defined Radios (SDRs) has introduced dense interference sources into the civil domain. Additionally, the rapid deployment of Autonomous Aerial Vehicles (AAVs) introduces dynamic interference, where high-mobility communication links and telemetry signals may overlap with navigation bands \cite{ref5}. These sources can saturate spectrum resources, degrading service in urban canyons and industrial zones \cite{ref8}.

The complexity of the spectral landscape has escalated significantly. While traditional mitigation techniques were designed for isolated interference sources, such as continuous waves or Gaussian noise, modern environments feature compound interference. Here, multiple diverse signal sources overlap in the time-frequency domain \cite{ref41}. For instance, a wide-band noise signal from a malfunctioning industrial device may mask the spectral footprint of a co-existing narrow-band signal, or high-power pulses from radar systems may interleave with frequency-sweeping chirps. This superposition results in an entangled signal structure where distinct physical characteristics interact non-linearly \cite{ref27}. As a result, conventional model-based detection methods, such as energy detectors or cyclostationary analysis, often fail to separate these mixed components, leading to high false alarm rates.

The core challenge in classifying such compound signals lies in the scale ambiguity of feature extraction and the masking effect of multi-modal features. Compound interference combines signal components with conflicting time-frequency granularities. Pulse-type interference manifests as micro-scale, transient bursts requiring high temporal resolution, whereas sweep-type interference, such as Linear Frequency Modulation (LFM), exhibits macro-scale, evolving global trends requiring broad spectral context. Existing Deep Learning (DL) approaches, primarily relying on standard Convolutional Neural Networks (CNNs) \cite{ref19} or Vision Transformers (ViTs) \cite{ref36} with fixed receptive fields, struggle to resolve these conflicting scales simultaneously \cite{ref6}. Moreover, relying solely on Time-Frequency Images (TFIs) can be unreliable; when strong wide-band noise is present, the fine-grained textural features of narrow-band components are often submerged, rendering single-domain classifiers ineffective\cite{ref6, ref27}.

Cognitive Radio (CR) offers a solution by advocating for intelligent systems capable of sensing and adapting to the electromagnetic environment. We propose a cognitive dual-stream framework, SKANet, designed to disentangle compound GNSS interference. We posit that robust classification requires two capabilities: adaptive feature extraction to handle multi-scale signal patterns, and multi-modal fusion to resolve domain-specific ambiguities. SKANet combines the temporal localization of TFIs with the statistical stability of Power Spectral Density (PSD). Central to our design is the integration of a Multi-Branch Selective Kernel (SK) mechanism \cite{ref28} with Asymmetric Convolution Blocks (ACBs) \cite{ref26}. This architecture allows the network to dynamically adjust its receptive fields, effectively focusing on different granularities to capture both transient pulses and continuous sweeps within an entangled signal mixture.

The main contributions of this paper are summarized as follows:
\begin{enumerate}
    \item We design a Multi-Branch SK-ACB module that integrates adaptive receptive field selection with asymmetric convolutions. This mechanism addresses the multi-scale issue of compound jamming, enabling robust feature extraction for both transient and continuous signal components.
      \item We construct a Cognitive Dual-Stream Architecture featuring a hierarchical asymmetric backbone. To mitigate the masking effect where wide-band noise obscures narrow-band features in single-domain representations, we design a deep semantic stream for TFIs and a lightweight statistical stream for PSD.
    \item We integrate a Squeeze-and-Excitation (SE) mechanism to adaptively recalibrate the importance of each modality.
    \item We conduct experiments on a dataset comprising nine types of compound interference signals. The results demonstrate that SKANet achieves an Overall Accuracy (OA) of 96.99\%, establishing a benchmark for compound jamming classification under low Jamming-to-Noise Ratio (JNR) regimes.
\end{enumerate}

The remainder of this paper is organized as follows. Section \ref{sec:related_work} reviews the existing work on GNSS interference suppression and deep learning-based perception techniques. Section \ref{sec:system_model} presents the mathematical models for the signal reception and the specific jamming primitives considered in this study. Section \ref{sec:signal_analysis} details the signal analysis methodology, explaining the generation of Time-Frequency Images and Power Spectral Density features. Section \ref{sec:methodology} elaborates on the proposed SKANet framework, providing a comprehensive description of the Multi-Branch SK-ACB module and the dual-stream architecture. Section \ref{sec:results} describes the experimental setup and provides a comprehensive analysis of the simulation results, including performance comparisons and ablation studies. Finally, Section \ref{sec:conclusion} concludes this paper.

\section{Related Work}
\label{sec:related_work}

The evolution of GNSS interference countermeasures has transitioned from model-driven signal processing to data-driven cognitive perception. This section reviews existing methodologies, analyzing their characteristics in handling compound interference and motivating the dual-stream design of SKANet.

\subsection{Model-Based Signal Processing Approaches}
Traditional interference mitigation relies on statistical properties and engineered features in the time-frequency domain. Early works utilized the Short-Time Fourier Transform (STFT) \cite{ref9} and Wigner-Ville Distribution (WVD) \cite{ref10} to detect spectral anomalies based on fixed thresholds. To mitigate stationary narrow-band signals, Adaptive Notch Filters (ANF) \cite{ref11, ref12} and Frequency Locked Loops (FLL) \cite{ref13} have been widely deployed. While computationally efficient, these methods often encounter a trade-off between tracking speed and stability, degrading when tracking highly dynamic signals with rapid frequency variations. Recent studies have explored multipolarization strategies \cite{ref45} and sensed equivalent bandwidth estimation \cite{ref42} to enhance suppression capabilities, yet these often require specialized antenna arrays or clean reference signals.

For non-stationary sweep interference, fractional domain methods have received attention. Sun \textit{et al.} exploited the energy aggregation property of the Fractional Fourier Transform (FrFT) to isolate chirp signals \cite{ref14, ref15}, while Luo \textit{et al.} utilized the sparsity of interferences in the Zak-transform domain \cite{ref16}. Other blind source separation techniques including Nonnegative Matrix Factorization (NMF) \cite{ref17} have also been explored. However, these transform-domain methods typically rely on a priori knowledge of signal parameters such as modulation rates or sparsity bases. Crucially, in compound interference scenarios where stochastic wideband noise overlaps with deterministic pulse trains, these model-based methods may struggle to disentangle the mixed signal components, leading to elevated false alarm rates.

\begin{figure*}[!t]
    \centering
    \includegraphics[width=6.0in]{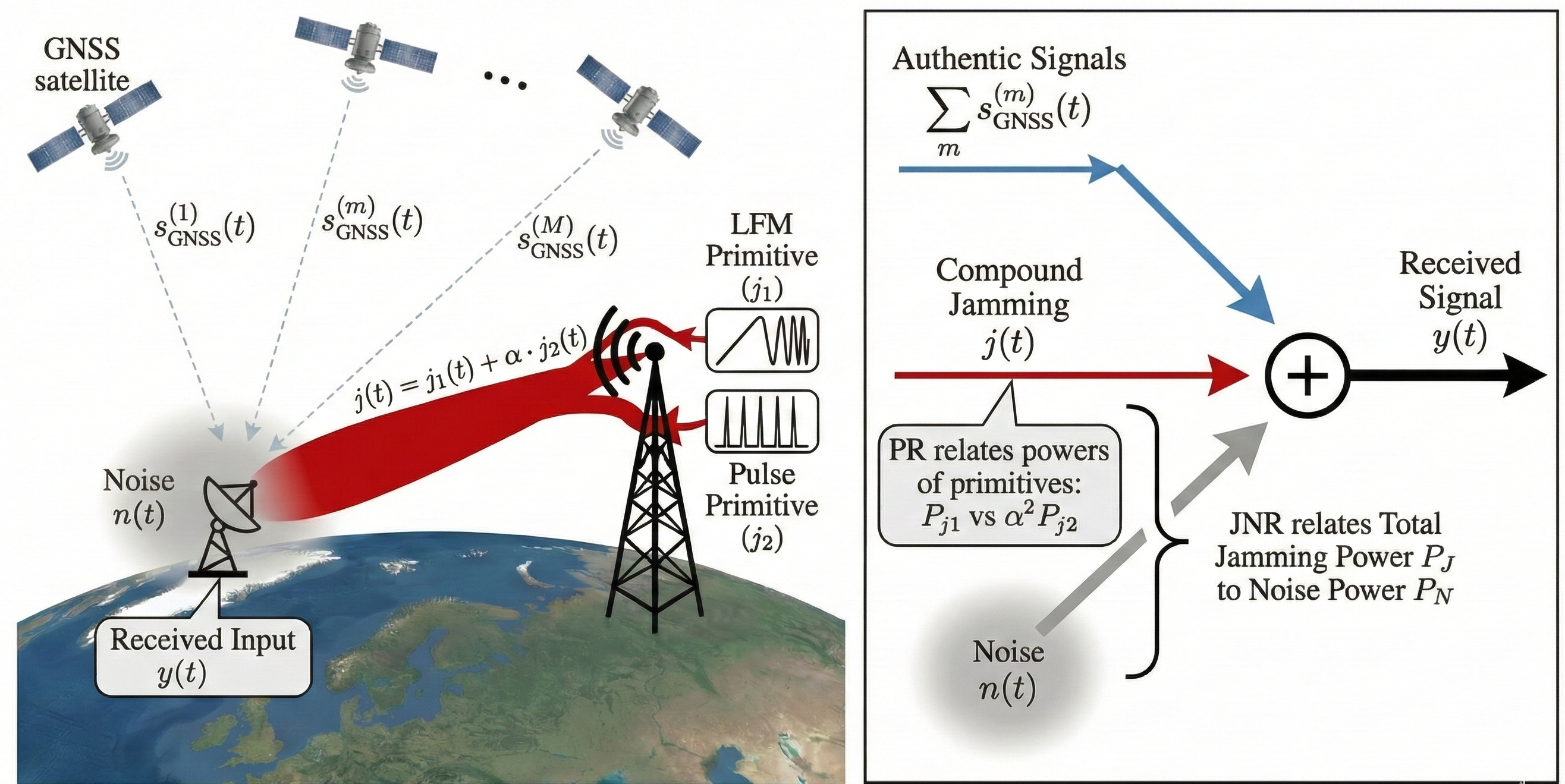}
    \caption{System model of the proposed GNSS interference scenario. 
(a) Physical layout of a GNSS receiver under compound jamming, where the source transmits superposed primitives (e.g., LFM and Pulse) mixed with a power ratio $\alpha$. 
(b) Signal superposition model at the receiver front-end. The received signal $y(t)$ is modeled as the summation of aggregate GNSS signals $\sum s_{\text{GNSS}}^{(m)}(t)$, compound jamming $j(t)$, and additive white Gaussian noise $n(t)$. Key parameters, including Power Ratio (PR) and Jamming-to-Noise Ratio (JNR), are defined based on the component power levels.}
    \label{fig:sys_model}
\end{figure*}

\begin{table}[!htpb]
\centering
\caption{Summary of Notations}
\label{tab:notation}
\renewcommand{\arraystretch}{1.3} 
\resizebox{\columnwidth}{!}{
\begin{tabular}{l|l} 
\toprule 
\textbf{Symbol} & \textbf{Description} \\ 
\midrule 
$a_c, b_c, c_c$ & Attention weights in SK module \\
$D$ & Set of dilation rates in SK-ACB module \\
$F_s$ & Sampling frequency \\
$\mathbf{F}_{final}$ & Final fused feature vector after SE block \\
$f_c, \phi$ & Carrier frequency offset and initial phase \\
$j(t)$ & Jamming signal \\
$\text{JNR}$ & Jamming-to-Noise Ratio \\
$K$ & Total number of interference classes \\
$\mathcal{L}(\Theta)$ & Cross-Entropy loss function \\
$N$ & Number of time-domain signal samples \\
$n(t)$ & Additive White Gaussian Noise (AWGN) \\
$P_J, \sigma_n^2$ & Jamming power and noise variance \\
$\text{PR}$ & Power Ratio for compound jamming \\
$\hat{P}_{\text{Welch}}(f)$ & Welch's Power Spectral Density (PSD) estimate \\
$S[m, k]$ & Logarithmic magnitude spectrogram (TFI) \\
$s_{\text{GNSS}}^{(m)}(t)$ & Signal from the $m$-th GNSS satellite \\
$T$ & Observation duration \\
$\tilde{\mathbf{U}}_m, \mathbf{V}$ & Intermediate and final feature maps in SK module \\
$X[m, k]$ & STFT coefficients at time frame $m$, freq. bin $k$ \\
$\mathbf{X}_{\text{TFI}}, \mathbf{X}_{\text{PSD}}$ & Input feature maps for TFI and PSD streams \\
$x(t)$ & Interference-plus-noise signal used for training \\
$y(t)$ & Received signal at the RF front-end \\
$\hat{\mathbf{y}}$ & Predicted probability distribution \\
$\alpha$ & Weighting coefficient for compound jamming \\
$\Theta$ & Set of all learnable network parameters \\
\bottomrule 
\end{tabular}%
}
\end{table}

\subsection{Deep Learning-Based Cognitive Perception}
The shift towards DL has enabled receivers to treat spectrum sensing as a visual recognition task. Early adopters demonstrated the efficacy of Support Vector Machines (SVM) \cite{ref18} and standard CNNs \cite{ref19, ref46} in classifying basic jamming primitives. To enhance feature robustness, recent advancements adapt computer vision architectures to the spectral domain. Zhang \textit{et al.} \cite{ref21} and Liu \textit{et al.} \cite{ref22} formulated interference detection as an object detection problem using YOLO-based frameworks. While effective for localized signals, bounding-box regression approaches face challenges when addressing distributed interference lacking distinct morphological boundaries. Alternatively, semantic segmentation models such as TF-Unet \cite{ref23} offer pixel-level granularity but often incur significant computational overhead for edge deployment.

Recent research focuses on strengthening backbone architectures. Jia \textit{et al.} proposed HI-CNN \cite{ref24} and lightweight few-shot classifiers \cite{ref31} to identify low-power interference, yet serial processing structures may introduce latency. Other architectures such as Dual Graph Convolutional Networks (GCN) \cite{ref32}, Hypergraph Networks \cite{ref47}, and GRU-based recurrent models \cite{ref33} investigate topological or temporal dependencies. Jiang \textit{et al.} introduced ACSNet \cite{ref25}, leveraging ACBs \cite{ref26} to enhance feature skeletonization. Concurrently, Xiao \textit{et al.} proposed TFPENet \cite{ref27} and fingerprint spectrum methods \cite{ref20}, utilizing dual-stream networks to fuse features. However, a notable challenge remains in many existing CNN-based and ViT \cite{ref36} approaches. These methods typically rely on static receptive fields that process the entire spectrogram with a fixed kernel size. This mechanism might not fully address the inherent scale ambiguity of compound jamming, particularly when micro-scale transient features and macro-scale spectral trends coexist. Furthermore, ViTs capable of modeling global dependencies may lack the inductive bias to efficiently capture high-frequency local textures in low-SNR environments \cite{ref44}, potentially affecting their efficacy for weak interference classification.

\subsection{Adaptive Receptive Fields and Modality Fusion}
Dynamic kernel selection mechanisms offer a potential solution to the multi-scale dilemma. The SK network \cite{ref28} introduces a mechanism where neurons adaptively adjust their receptive field sizes based on input stimuli. This concept has demonstrated potential in remote sensing \cite{ref29} and modulation recognition \cite{ref30} but remains limited in GNSS anti-jamming.

SKANet addresses this gap by applying the SK mechanism specifically to the spectral characteristics of compound jamming. Unlike ACSNet \cite{ref25} or TFPENet \cite{ref27} which employ fixed feature extractors, SKANet combines multi-modal fusion with adaptive receptive fields. By integrating Multi-Branch SK units with ACBs, our framework creates a cognitive cycle that dynamically modulates attention between transient temporal features and continuous spectral trends. This design allows the network to function as an adaptive filter bank, resolving the scale conflict inherent in entangled signal mixtures while maintaining the robustness provided by dual-domain analysis using both TFI and PSD data.

For ease of reference, the key mathematical notations and symbols used throughout this paper are summarized in Table \ref{tab:notation}.


\section{System Model}
\label{sec:system_model}

This section establishes the theoretical framework for the GNSS interference scenarios considered in this study. We detail the signal reception model, the mathematical formulations of individual jamming primitives, and the superposition mechanism for compound interference.

\subsection{Signal Reception Model}
We consider a typical GNSS interference scenario as illustrated in Fig. \ref{fig:sys_model}, where a receiver is subjected to malicious jamming attacks. The received signal $y(t)$ at the RF front-end, after down-conversion to baseband, can be modeled as the superposition of authentic satellite signals, jamming interference, and noise as follows:
\begin{equation}
    y(t) = \sum_{m=1}^{M} s_{\text{GNSS}}^{(m)}(t) + j(t) + n(t),
    \label{eq:rx_model}
\end{equation}
where $M$ denotes the number of visible GNSS satellites, $s_{\text{GNSS}}^{(m)}(t)$ represents the signal from the $m$-th satellite, $j(t)$ denotes the jamming signal, and $n(t)$ is complex Additive White Gaussian Noise (AWGN) with zero mean and variance $\sigma_n^2$\cite{ref6, ref9}.

In the context of jamming detection and classification, the jamming power $P_J$ is typically much higher than the thermal noise power $P_N$ and the authentic signal power $P_S$ (i.e., $P_J \gg P_S$). Thus, the authentic GNSS signals are often submerged beneath the jamming and noise floor. For training the classification model, we focus on the interference-plus-noise component given as follows:
\begin{equation}
    x(t) = j(t) + n(t).
\end{equation}
Let $F_s$ denote the sampling frequency and $T$ denote the observation duration. The discrete-time signal is given by $x[n] = x(n/F_s)$ for $n = 0, \dots, N-1$.

\subsection{Single Jamming Primitives}
\begin{figure}[!t]
    \centering
    \includegraphics[width=3.4in]{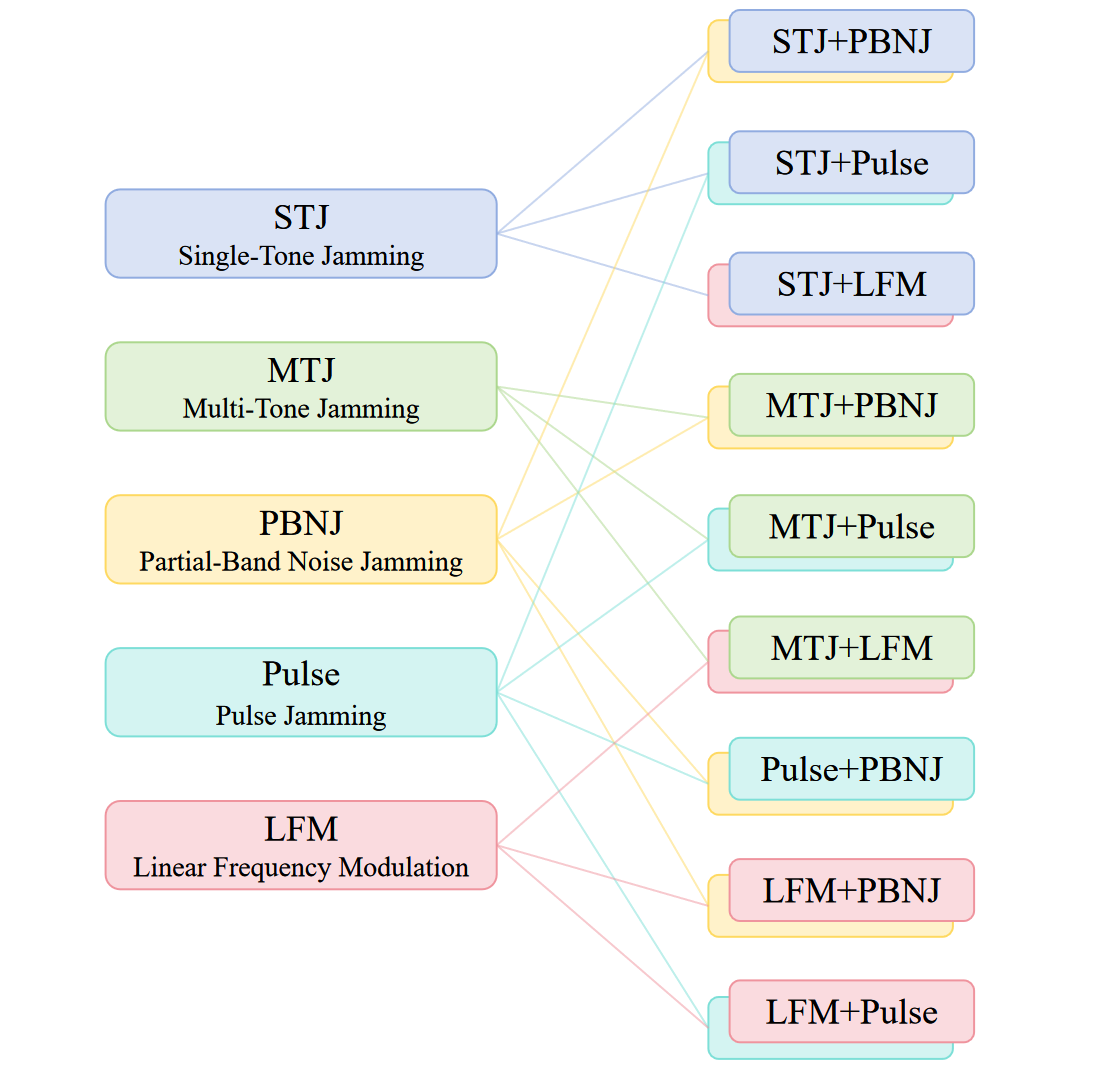}
   \caption{The hierarchical taxonomy of GNSS interference types considered in the proposed framework. The model categorizes interference into five fundamental single primitives (STJ, MTJ, LFM, Pulse, PBNJ) and their compound variations formed by linear superposition. This hierarchy serves as the basis for the dataset generation and the nine-class classification task, distinguishing between isolated jamming sources and complex entangled signal mixtures.}
    \label{fig:jam_taxonomy}
\end{figure}

To construct a robust classification framework, we first define the fundamental single-source jamming types (primitives).  Based on the taxonomy shown in Fig. \ref{fig:jam_taxonomy}, we consider five distinct categories. Each type exhibits unique time-frequency characteristics and impacts the receiver tracking loops in different ways.

\subsubsection{Single-Tone Jamming (STJ)}
STJ, also known as Continuous Wave (CW) interference, is common due to its simplicity. It is characterized by a constant frequency and amplitude, manifesting as a sharp spectral peak in the frequency domain. STJ primarily affects the receiver by saturating the Automatic Gain Control (AGC) or by capturing the Phase Locked Loop (PLL), causing the receiver to lose lock on the authentic satellite signal. The mathematical expression for the complex baseband STJ signal is given by
\begin{equation}
    j_{\text{STJ}}(t) = \sqrt{P_J} e^{j(2\pi f_c t + \phi)},
\end{equation}
where $P_J$ denotes the jamming power, $f_c$ represents the carrier frequency offset relative to the GNSS center frequency, and $\phi$ is the initial phase.

\subsubsection{Multi-Tone Jamming (MTJ)}
MTJ comprises a superposition of multiple independent single-tone signals. Unlike STJ, MTJ disperses its energy across several discrete frequency points, aiming to simultaneously disrupt multiple sub-bands or signal components. This structure complicates mitigation, as adaptive notch filters must track and suppress multiple peaks. The MTJ signal is formulated as:
\begin{equation}
    j_{\text{MTJ}}(t) = \sum_{k=1}^{K} \sqrt{\frac{P_J}{K}} e^{j(2\pi f_k t + \phi_k)},
\end{equation}
where $K$ represents the number of tones. $f_k$ and $\phi_k$ denote the frequency and initial phase of the $k$-th tone, respectively. The spacing between adjacent tones determines the spectral density of the attack.

\subsubsection{Linear Frequency Modulation (LFM)}
LFM, widely referred to as chirp jamming, is a sweep interference where the instantaneous frequency varies linearly with time. LFM is particularly effective against spread spectrum systems as it sweeps across the entire signal bandwidth, periodically degrading the correlation output and maximizing the Bit Error Rate (BER)\cite{ref14, ref15}. The complex baseband model of LFM is expressed as:
\begin{equation}
    j_{\text{LFM}}(t) = \sqrt{P_J} e^{j 2\pi (f_0 t + \frac{k}{2} t^2)},
\end{equation}
where $f_0$ is the starting frequency, and $k = B/T_{\text{sw}}$ denotes the frequency modulation slope (chirp rate), determined by the sweep bandwidth $B$ and sweep period $T_{\text{sw}}$.

\subsubsection{Pulse Jamming}
Pulse jamming is a time-domain non-stationary interference characterized by high-power bursts of energy emitted at specific intervals. It creates a high Peak-to-Average Power Ratio (PAPR), which can cause severe quantization errors in the Analog-to-Digital Converter (ADC) and disrupt the receiver's AGC synchronization\cite{ref9}. The pulse signal is modeled as a gated sinusoid as follows:
\begin{equation}
    j_{\text{Pulse}}(t) = p(t) \cdot \sqrt{P_J} e^{j 2\pi f_c t},
\end{equation}
where $p(t)$ is a rectangular pulse train function defined by the Pulse Repetition Interval (PRI) and Pulse Width (PW). The duty cycle determines the temporal density of the attack.

\subsubsection{Partial-Band Noise Jamming (PBNJ)}
PBNJ is a stochastic interference type that distributes noise energy within a specific bandwidth $B_J$, rather than across the entire spectrum like broadband thermal noise. It is typically generated by passing white Gaussian noise through a band-limited filter. PBNJ effectively raises the noise floor in the target band, thereby reducing the Signal-to-Noise Ratio (SNR) and masking the weak GNSS signals. Mathematically, it is described as:
\begin{equation}
    j_{\text{PBNJ}}(t) = \left( h_{\text{LPF}}(t) * w(t) \right) e^{j 2\pi f_c t},
\end{equation}
where the operator $\ast$ denotes convolution, $w(t)$ is standard complex Gaussian white noise, and $h_{\text{LPF}}(t)$ represents the impulse response of a shaping filter (e.g., a Butterworth low-pass filter) with a cutoff frequency corresponding to the jamming bandwidth.

\subsection{Compound Jamming Model}
In complex electromagnetic environments, multiple interference sources often operate simultaneously, resulting in a compound signal structure. This work defines compound interference as the linear superposition of two distinct single jamming primitives.

Let $j_1(t)$ and $j_2(t)$ denote two independent, unit-power jamming components selected from the primitive set defined above\cite{ref27, ref41}. The compound interference signal $j_{\text{mix}}(t)$ is formally modeled as:
\begin{equation}
    j_{\text{mix}}(t) = \frac{1}{\gamma} \left( j_1(t) + \alpha \cdot j_2(t) \right),
    \label{eq:compound}
\end{equation}
where $\gamma$ is a normalization factor constraining the total power, and $\alpha$ is a weighting coefficient that determines the relative intensity between the two components.

This study systematically considers compound jamming classes formed by pairing different primitives. Specifically, these combinations include the mixture of STJ with LFM, Pulse jamming, and PBNJ. Furthermore, MTJ is superimposed with LFM, Pulse, and PBNJ respectively. The model also accounts for combinations of dynamic interference types, namely LFM mixed with Pulse and LFM mixed with PBNJ, as well as the combination of Pulse and PBNJ. These diverse scenarios represent realistic electronic warfare environments where narrow-band, sweep, and wide-band interferences coexist and overlap in the time-frequency domain.

To capture the varying dominance between interference sources, we utilize the Power Ratio (PR)~\cite{ref27}. The PR quantifies the power difference between the secondary component relative to the primary component and is expressed in decibels (dB) as:
\begin{equation}
    PR_{\text{dB}} = 10 \log_{10}(\alpha^2) = 20 \log_{10}(\alpha).
    \label{eq:PR}
\end{equation}
This formulation ensures that the model can represent scenarios ranging from equal-power superposition to cases where one jammer is significantly subordinate to the other~\cite{ref25}. The overall severity of the interference relative to the environmental noise is quantified by the JNR, defined as $\text{JNR}_{\text{dB}} = 10 \log_{10} (P_J/\sigma_n^2)$~\cite{ref6, ref9}.


\section{Signal Analysis}
\label{sec:signal_analysis}

Effective classification of GNSS interference relies on the extraction of discriminative features from the raw signal. Since jamming signals exhibit diverse characteristics—ranging from stationary narrow-band tones to non-stationary wide-band sweeps—a single domain representation is often insufficient. Therefore, we employ a dual-domain analysis approach combining STFT for time-frequency localization and PSD for energy distribution analysis.

\begin{figure}[!t]
\centering
\subfloat[STJ+LFM]{\includegraphics[width=0.29\linewidth]{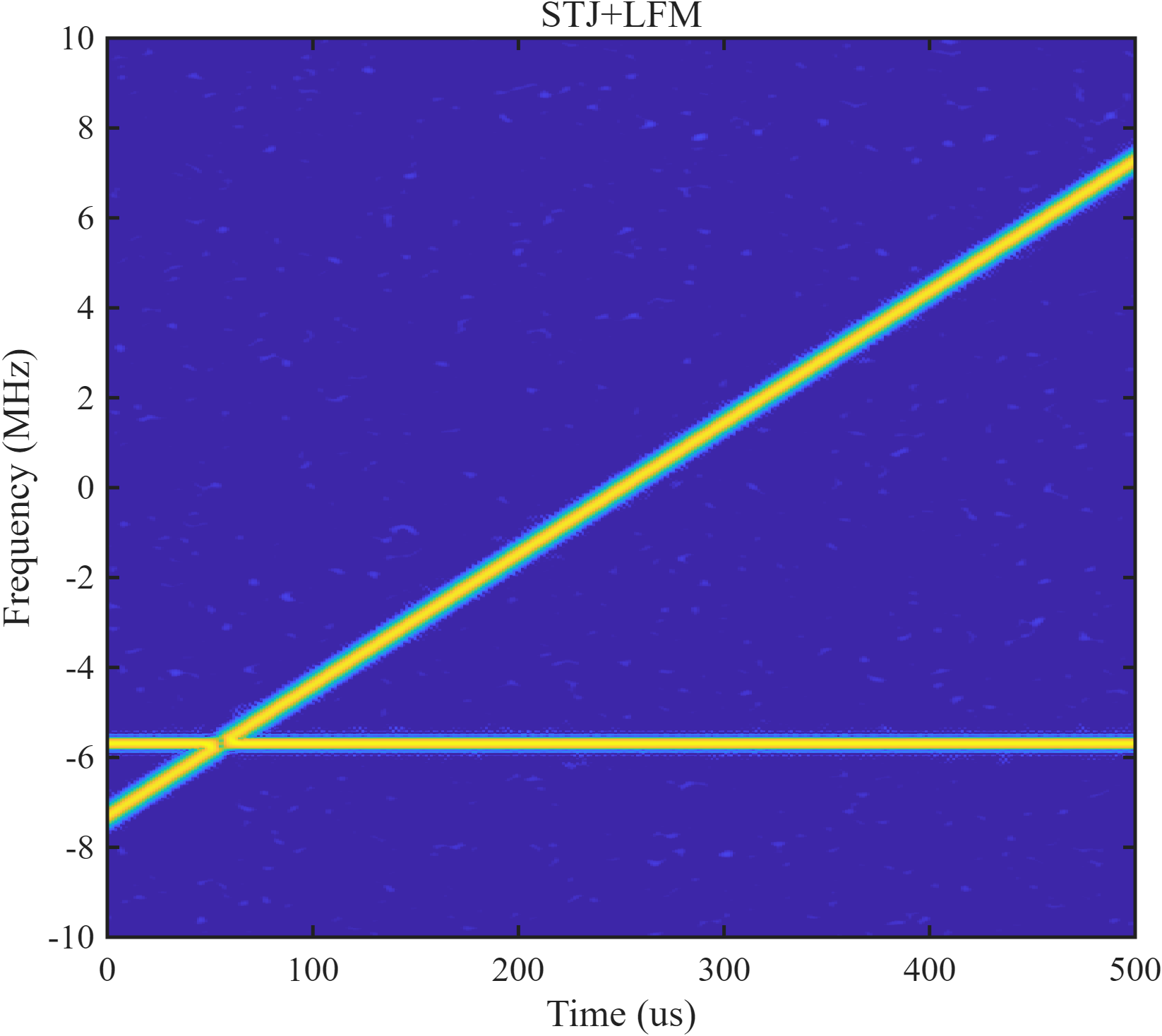}
\label{fig:stft_stj_lfm}} \hfil
\subfloat[STJ+Pulse]{\includegraphics[width=0.29\linewidth]{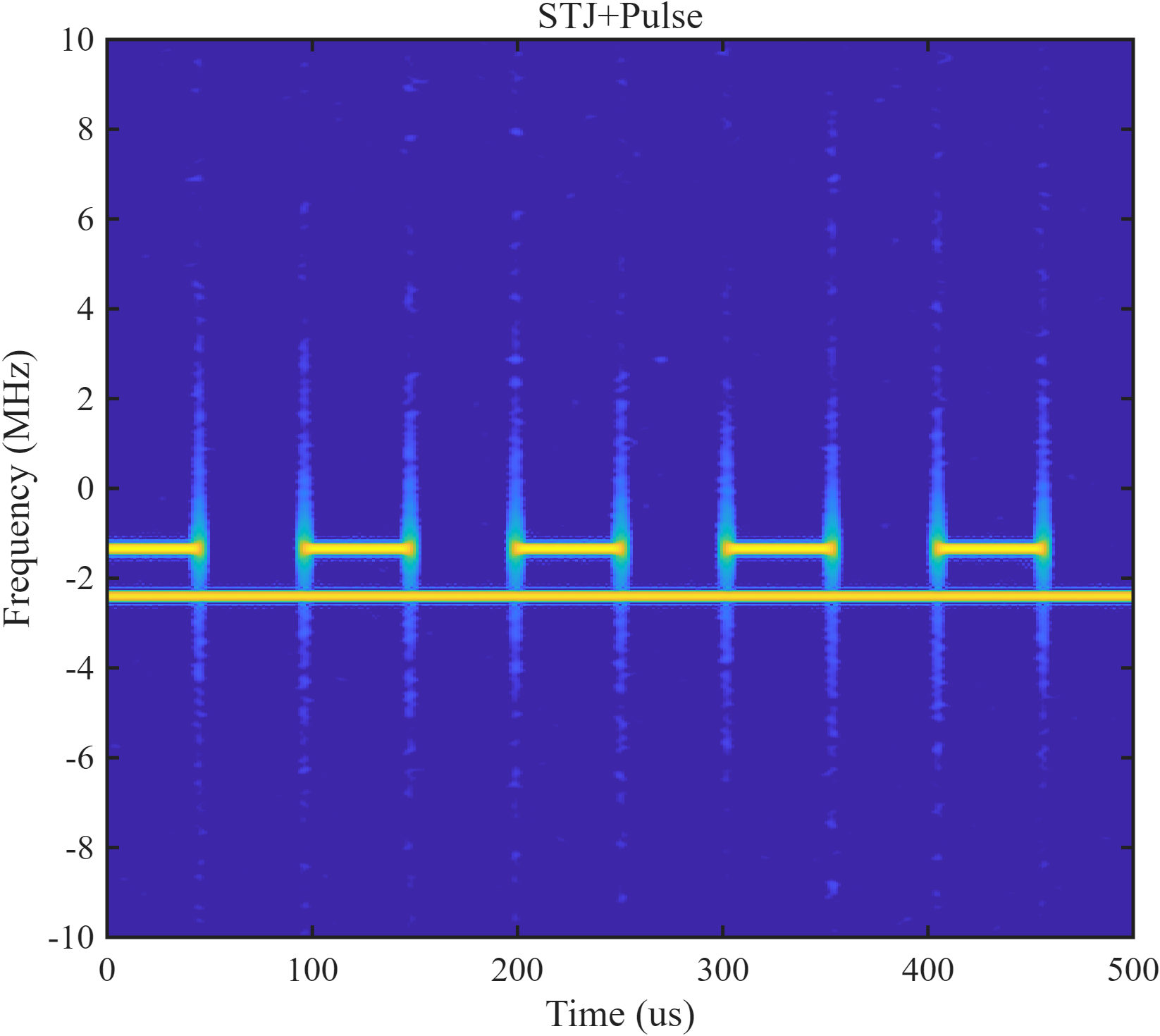}
\label{fig:stft_stj_pulse}} \hfil
\subfloat[STJ+PBNJ]{\includegraphics[width=0.29\linewidth]{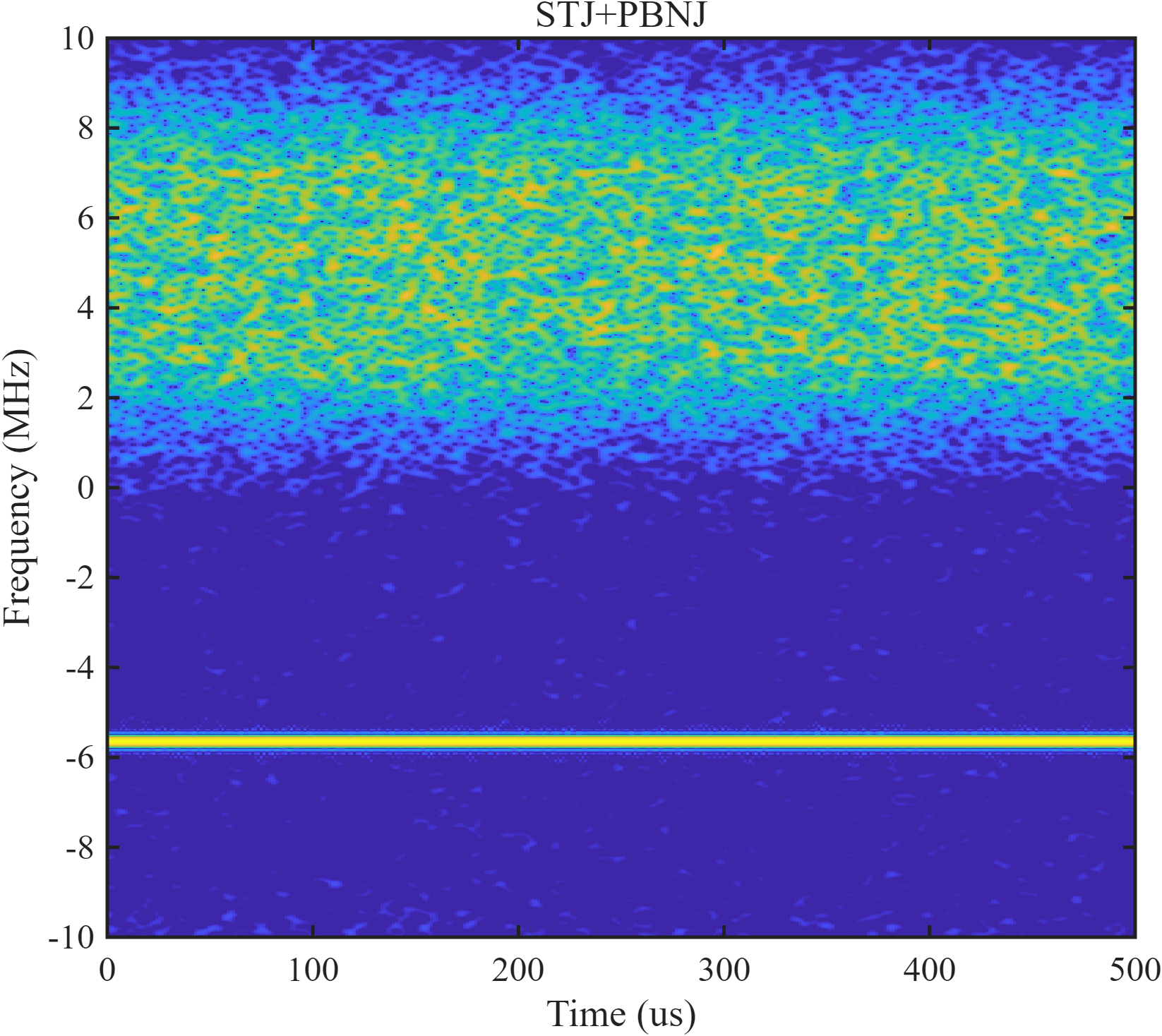}
\label{fig:stft_stj_pbnj}} \\

\vspace{-0.5em}
\subfloat[MTJ+LFM]{\includegraphics[width=0.29\linewidth]{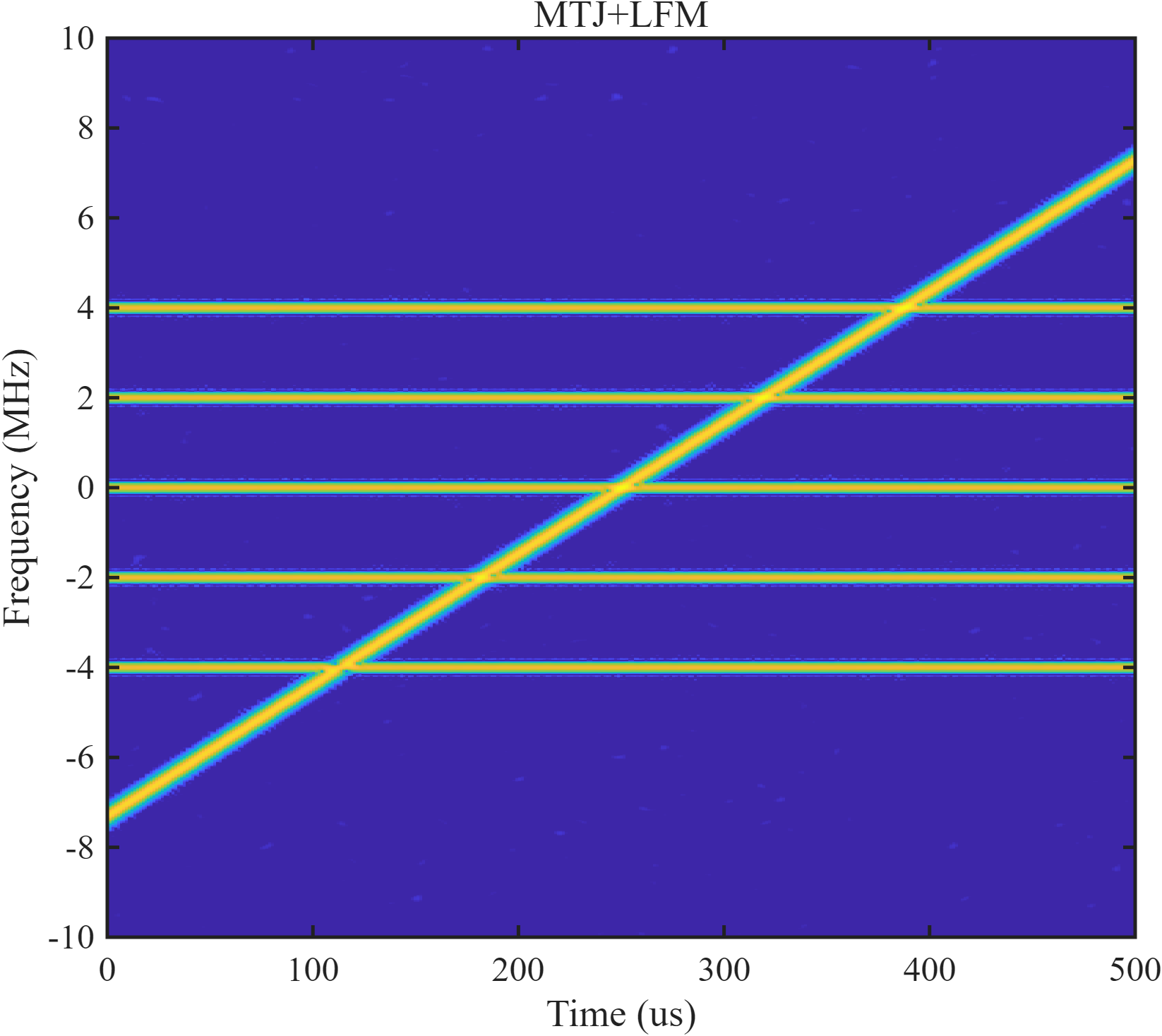}
\label{fig:stft_mtj_lfm}} \hfil
\subfloat[MTJ+Pulse]{\includegraphics[width=0.29\linewidth]{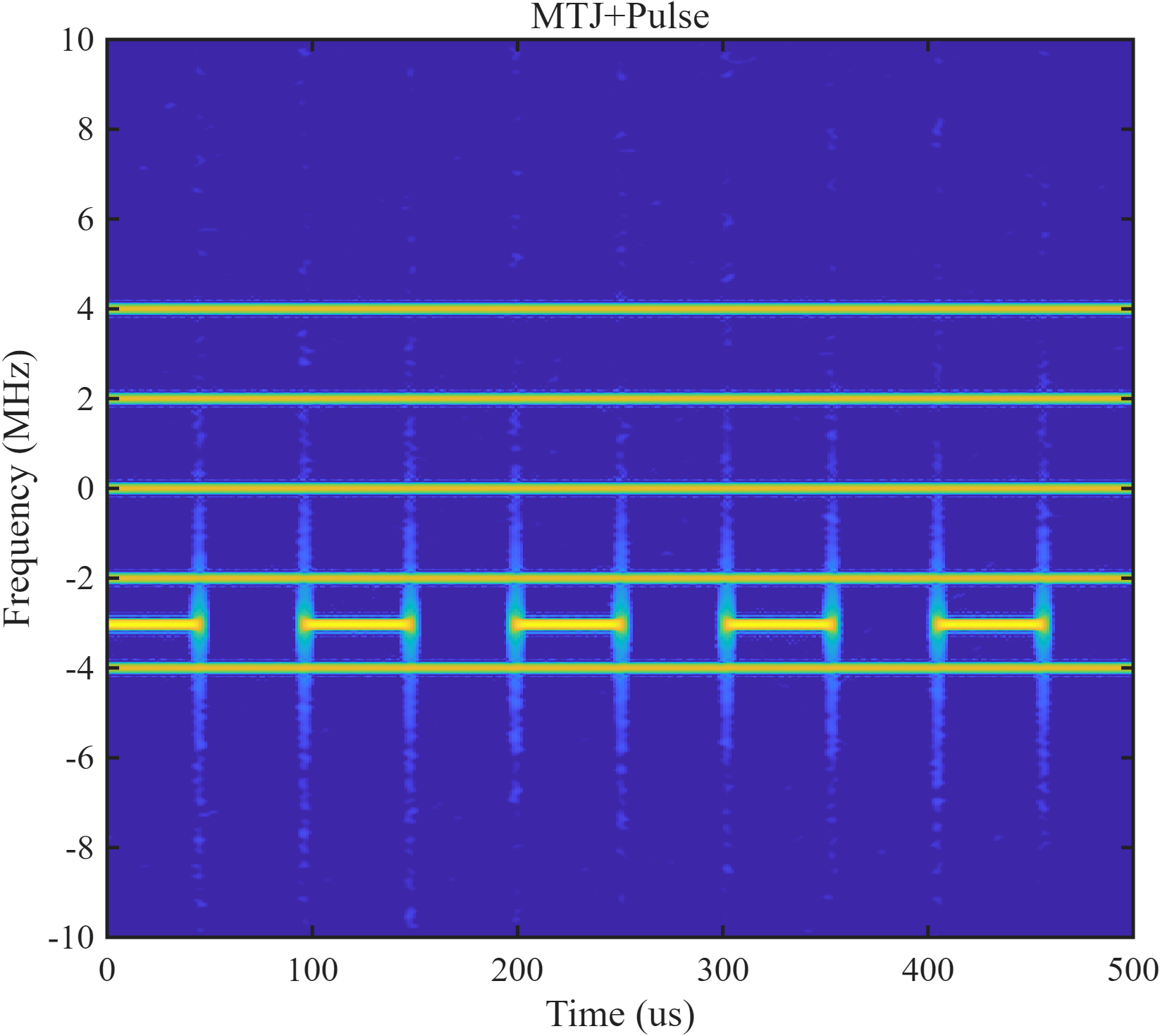}
\label{fig:stft_mtj_pulse}} \hfil
\subfloat[MTJ+PBNJ]{\includegraphics[width=0.29\linewidth]{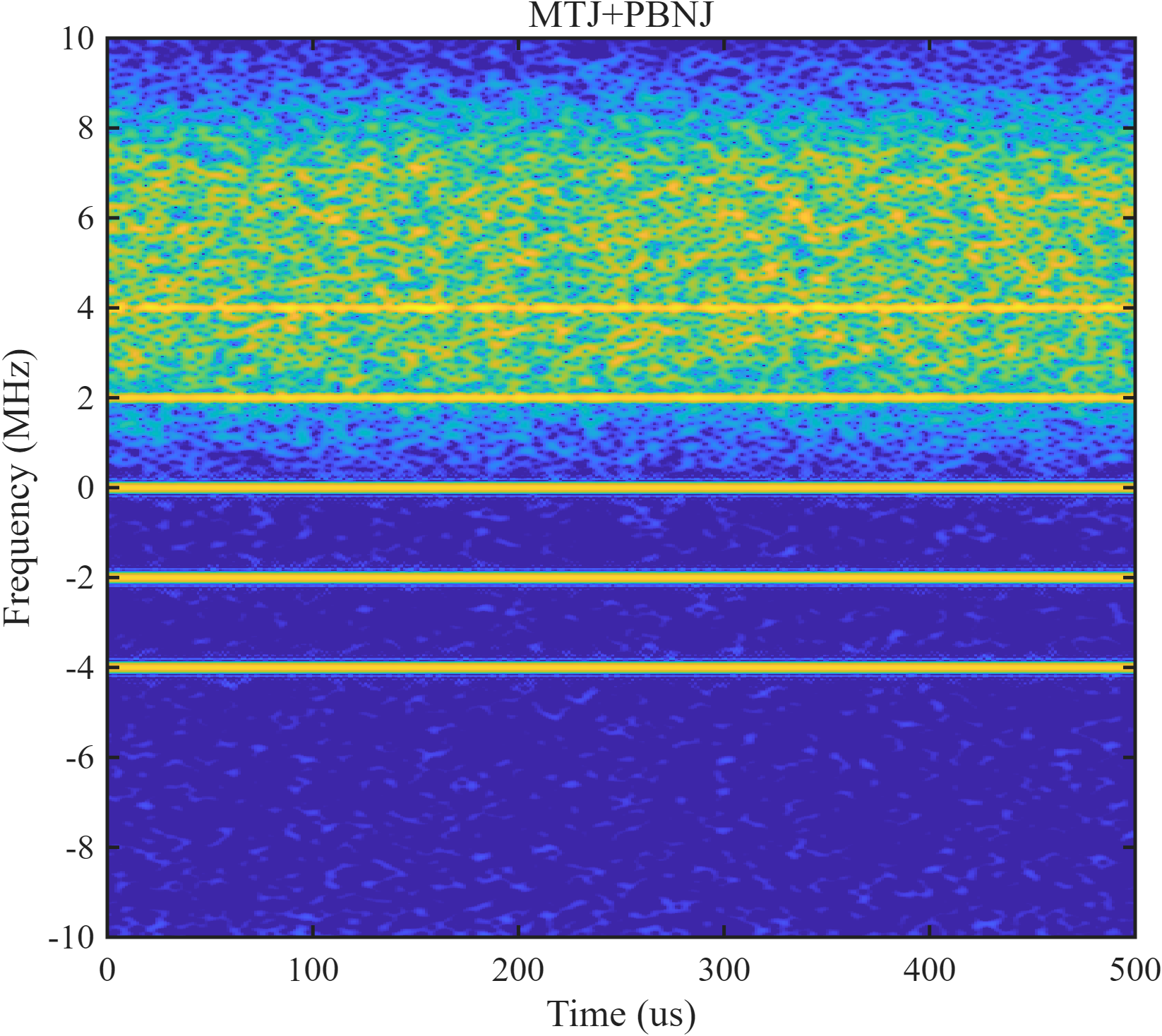}
\label{fig:stft_mtj_pbnj}} \\

\vspace{-0.5em}
\subfloat[LFM+Pulse]{\includegraphics[width=0.29\linewidth]{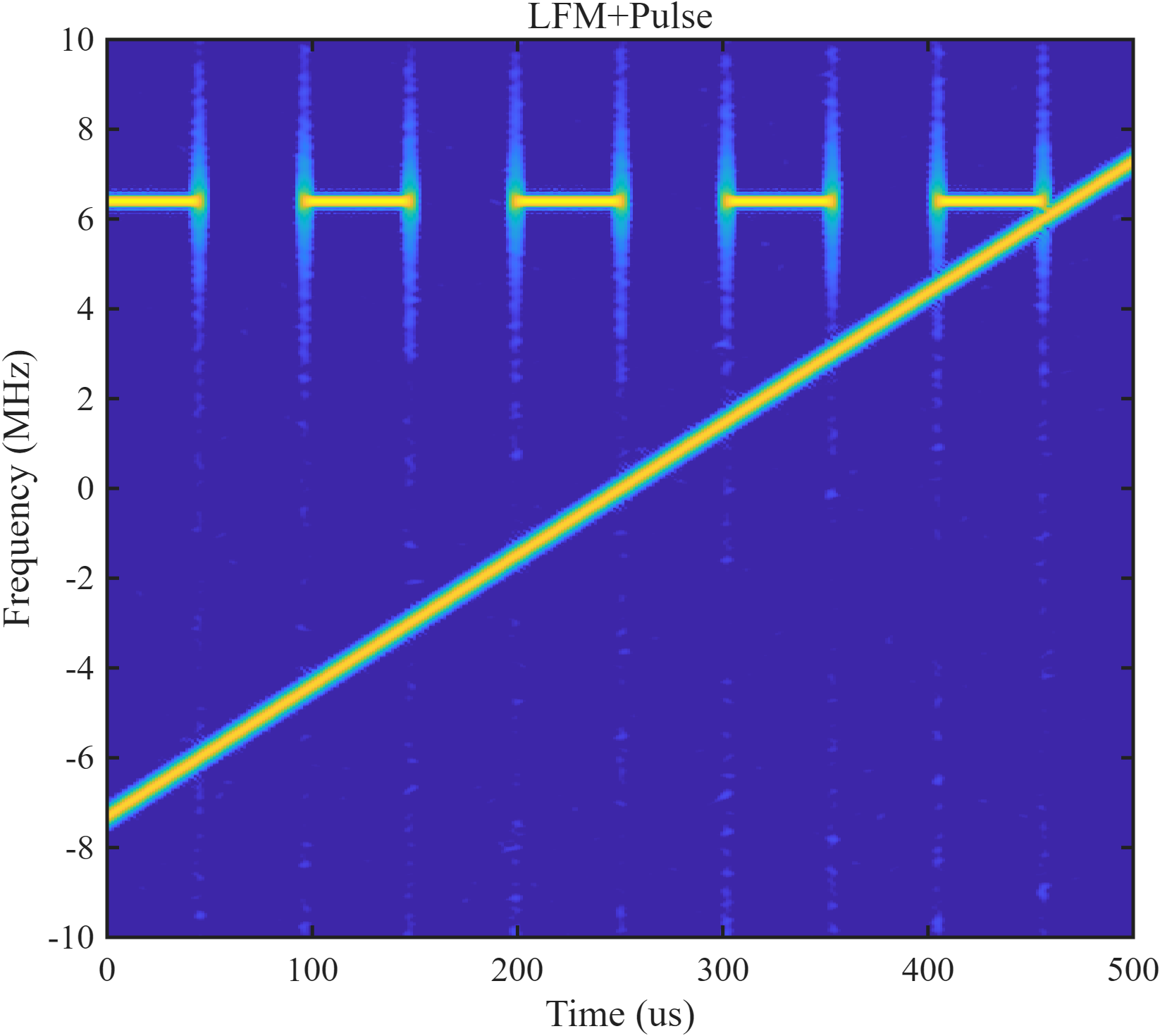}
\label{fig:stft_lfm_pulse}} \hfil
\subfloat[LFM+PBNJ]{\includegraphics[width=0.29\linewidth]{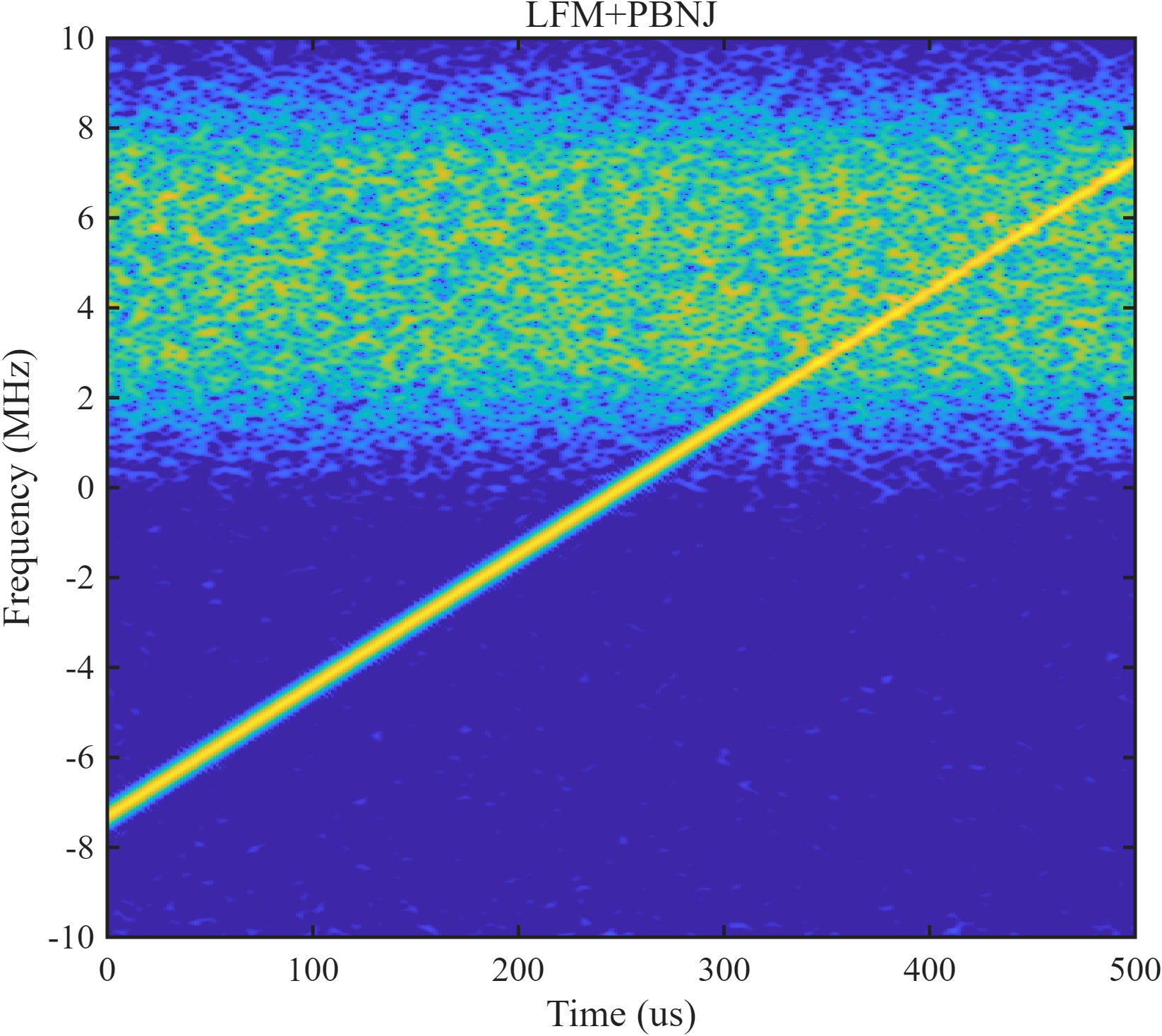}
\label{fig:stft_lfm_pbnj}} \hfil
\subfloat[Pulse+PBNJ]{\includegraphics[width=0.29\linewidth]{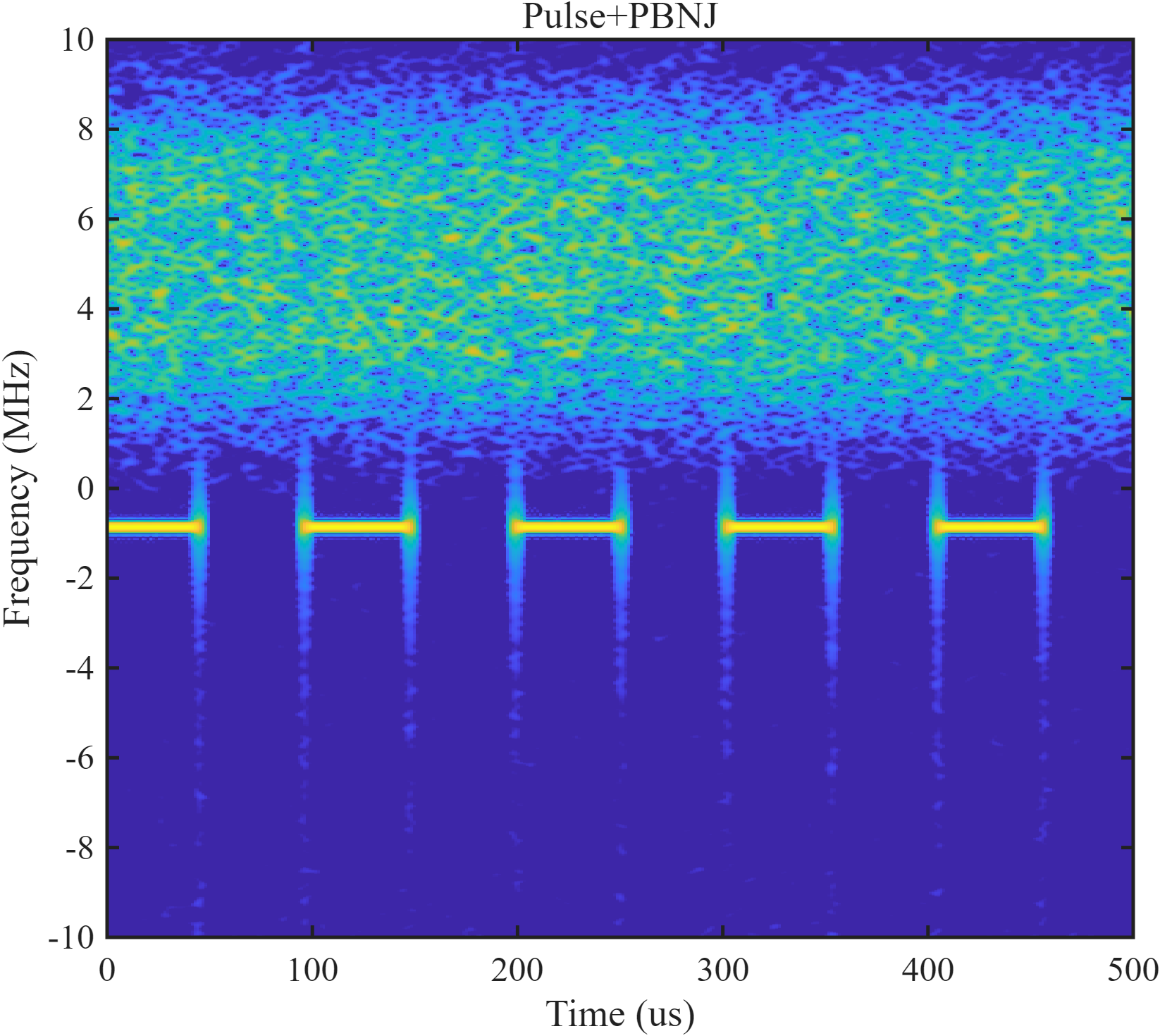}
\label{fig:stft_pulse_pbnj}}

\caption{Time-Frequency Images (TFIs) of the 9 compound interference patterns generated via STFT. The 3$\times$3 grid layout illustrates the distinct spectral overlap characteristics of different jamming combinations within the limited GNSS bandwidth.}
\label{fig:compound_stft}
\end{figure}

\subsection{Short-Time Fourier Transform (STFT)}
The STFT is fundamental for analyzing non-stationary signals, such as LFM and Pulse jamming, where frequency components vary over time. It maps the one-dimensional time-domain signal $x[n]$ into a two-dimensional time-frequency matrix.

Mathematically, the STFT of the discrete signal $x[n]$ is computed by sliding a window function $w[n]$ along the signal and performing the Discrete Fourier Transform (DFT) on each segment\cite{ref25}. The STFT matrix $X[m, k]$ is defined as:
\begin{equation}
    X[m, k] = \sum_{n=0}^{N_w-1} x[n + mH] w[n] e^{-j \frac{2\pi}{N_{\text{fft}}} k n}.
\end{equation}
In this formulation, $m$ and $k$ denote the discrete time frame index and frequency bin index, respectively. The parameter $N_w$ represents the length of the window function, while $H$ defines the hop size (or step size) between consecutive sliding windows. Additionally, $N_{\text{fft}}$ specifies the total number of points used for the DFT.

To minimize spectral leakage, we employ the Hann window function\cite{ref9}, defined as:
\begin{equation}
    w[n] = 0.5 \left( 1 - \cos\left(\frac{2\pi n}{N_w - 1}\right) \right), \quad 0 \leq n \leq N_w-1.
\end{equation}

For neural network inference, the phase information is typically discarded in favor of the magnitude spectrogram. To compress the high dynamic range of jamming signals (where $P_J \gg P_N$) and enhance the visibility of weak interference structures, we apply a logarithmic transformation to the magnitude squared of the STFT coefficients defined as follows:
\begin{equation}
    S[m, k] = 10 \log_{10} \left( |X[m, k]|^2 + \epsilon \right),
\end{equation}
where $\epsilon$ is a small constant to prevent numerical instability. The resulting spectrogram $S[m, k]$ visualizes the energy distribution over time and frequency, providing key features for identifying transient interference patterns.

\subsection{Power Spectral Density (PSD)}
While STFT captures temporal variations \cite{ref25}, it is computationally intensive and may suffer from resolution trade-offs. The PSD provides a robust statistical estimate of the signal's power distribution across frequencies, which is particularly effective for characterizing stationary interferences including STJ, MTJ, and PBNJ. We utilize Welch's method to estimate the PSD, as it reduces the variance of the periodogram through averaging. The signal $x[n]$ of length $N$ is divided into $L$ overlapping segments $x_i[n]$ ($i = 1, \dots, L$). For each segment of length $M$, a modified periodogram is calculated using a window function $w_{PSD}[n]$ (typically a Hamming window). The final Welch PSD estimate is obtained by averaging these periodograms, ensuring an asymptotically unbiased and statistically stable spectral representation. The calculation process is formulated as follows:
\begin{align}
    & U = \frac{1}{M} \sum_{n=0}^{M-1} |w_{PSD}[n]|^2, \label{eq:welch_norm} \\
    & \hat{P}_i(f) = \frac{1}{M U} \left| \sum_{n=0}^{M-1} x_i[n] w_{PSD}[n] e^{-j 2\pi f n} \right|^2, \label{eq:welch_peridogram} \\
    & \hat{P}_{\text{Welch}}(f) = \frac{1}{L} \sum_{i=1}^{L} \hat{P}_i(f). \label{eq:welch_final}
\end{align}
By utilizing Welch's method with a Hamming window and a 50\% overlap, we obtain a smooth spectral representation. This overlap ratio is a standard configuration in spectral estimation to balance the trade-off between variance reduction and the correlation of adjacent segments \cite{proakis2007digital}. Consequently, the derived spectral profile serves as a complementary feature input to the neural network. It aggregates global energy statistics to distinguish interferences that may exhibit ambiguous morphological boundaries in the time-frequency domain but possess distinct power spectral footprints in the frequency domain.

\begin{figure}[!t]
\centering
\subfloat[STJ+LFM]{\includegraphics[width=0.29\linewidth]{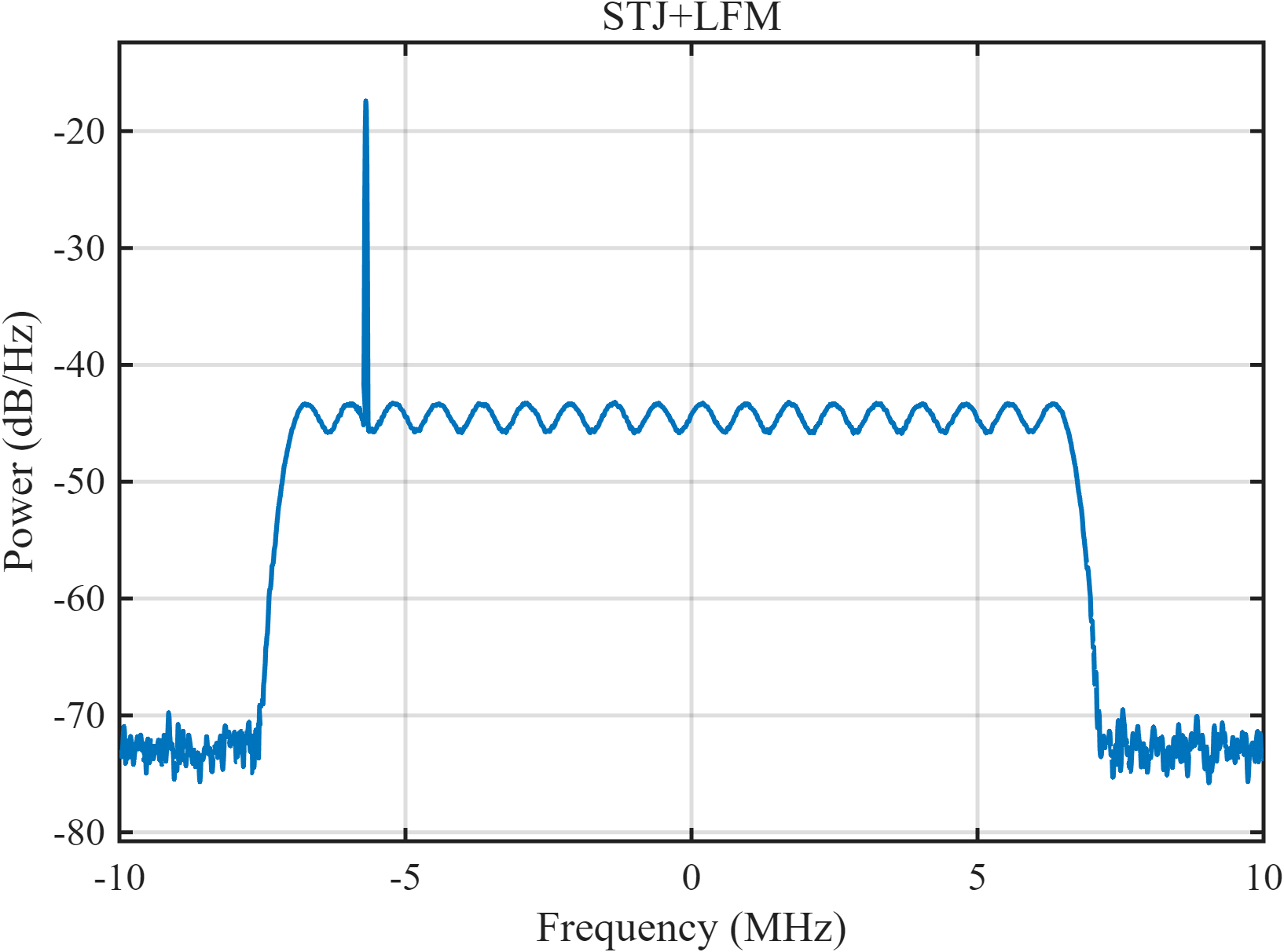}
\label{fig:psd_stj_lfm}} \hfil
\subfloat[STJ+Pulse]{\includegraphics[width=0.29\linewidth]{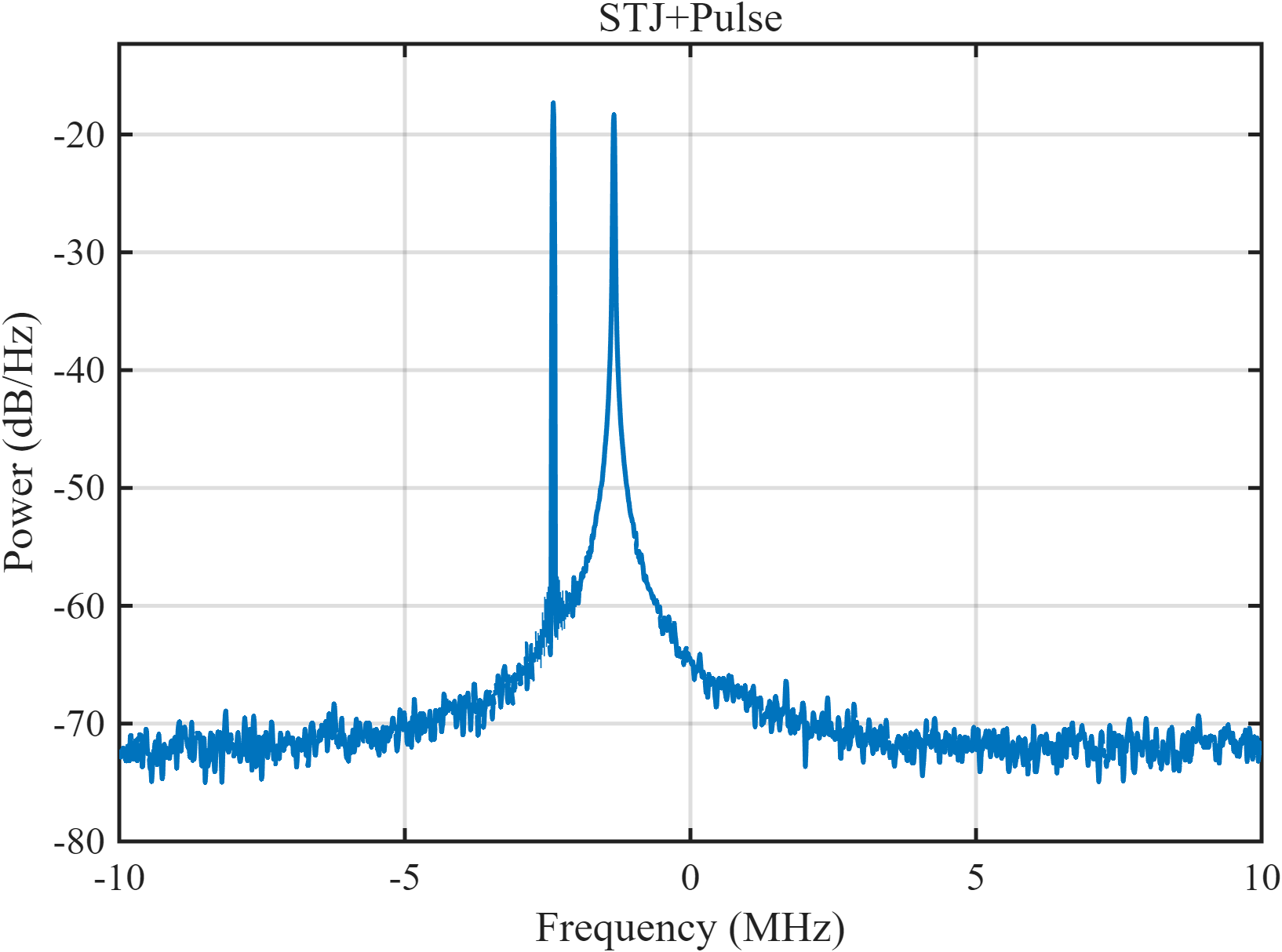}
\label{fig:psd_stj_pulse}} \hfil
\subfloat[STJ+PBNJ]{\includegraphics[width=0.29\linewidth]{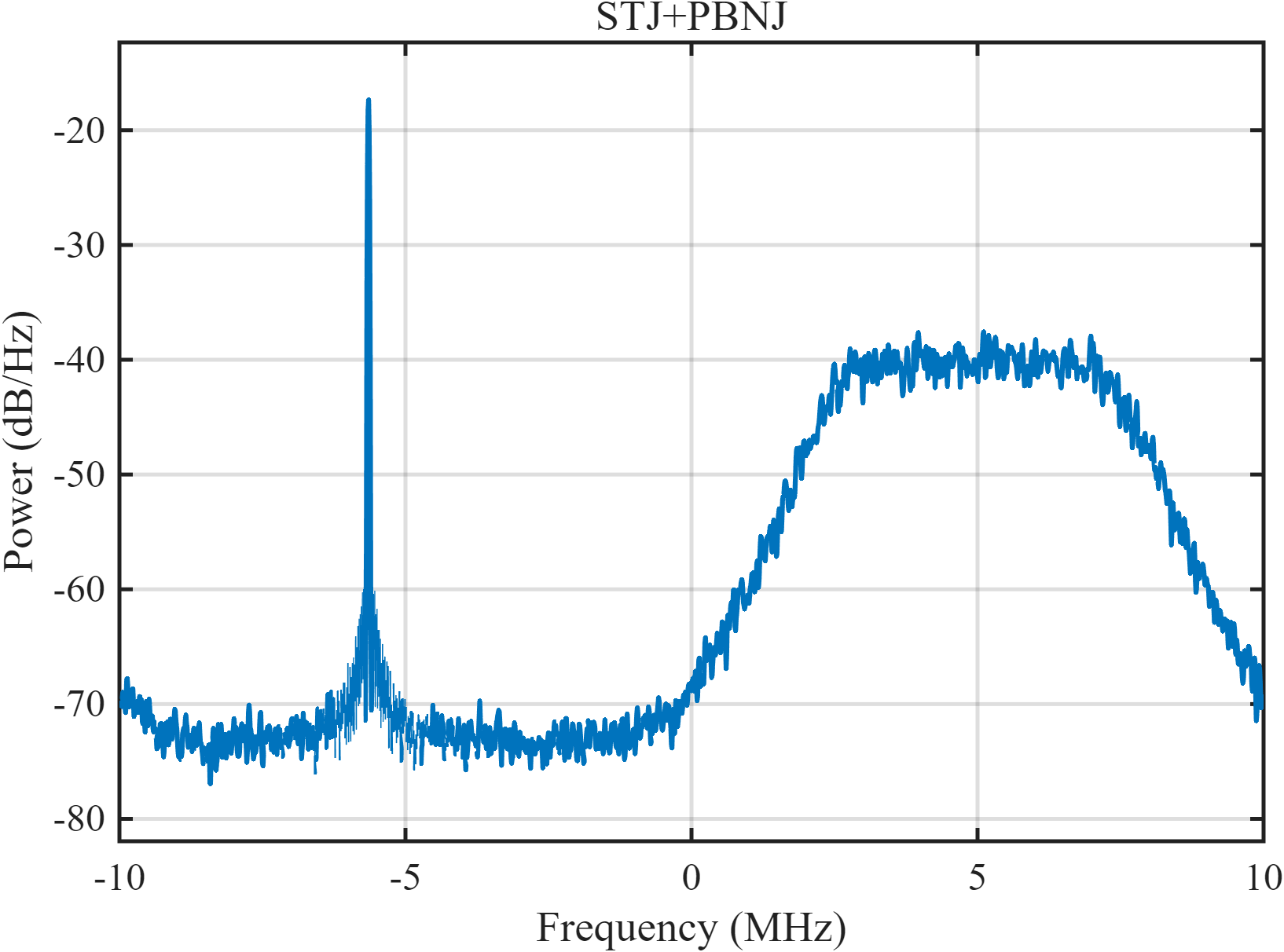}
\label{fig:psd_stj_pbnj}} \\

\vspace{1em}
\subfloat[MTJ+LFM]{\includegraphics[width=0.29\linewidth]{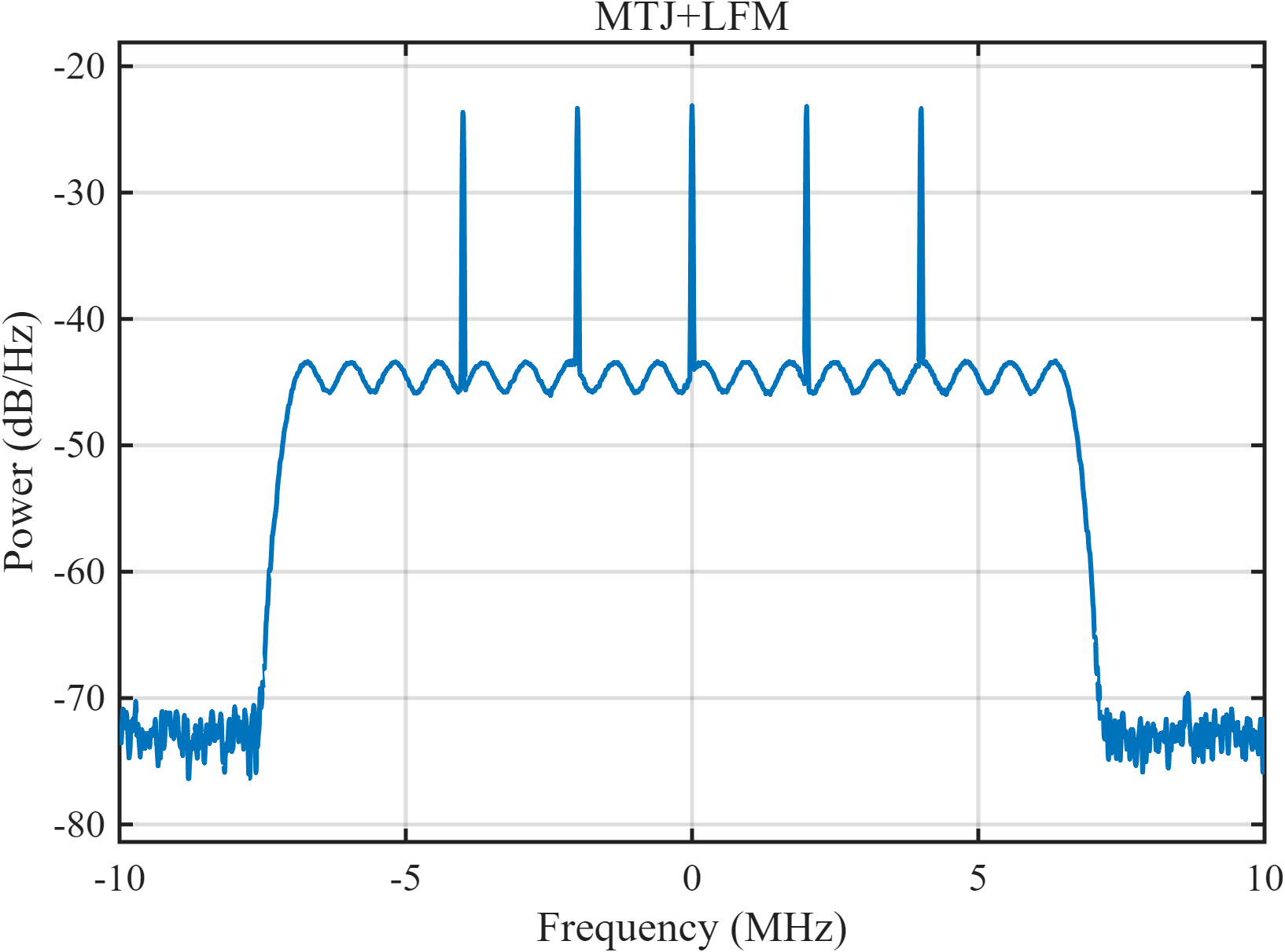}
\label{fig:psd_mtj_lfm}} \hfil
\subfloat[MTJ+Pulse]{\includegraphics[width=0.29\linewidth]{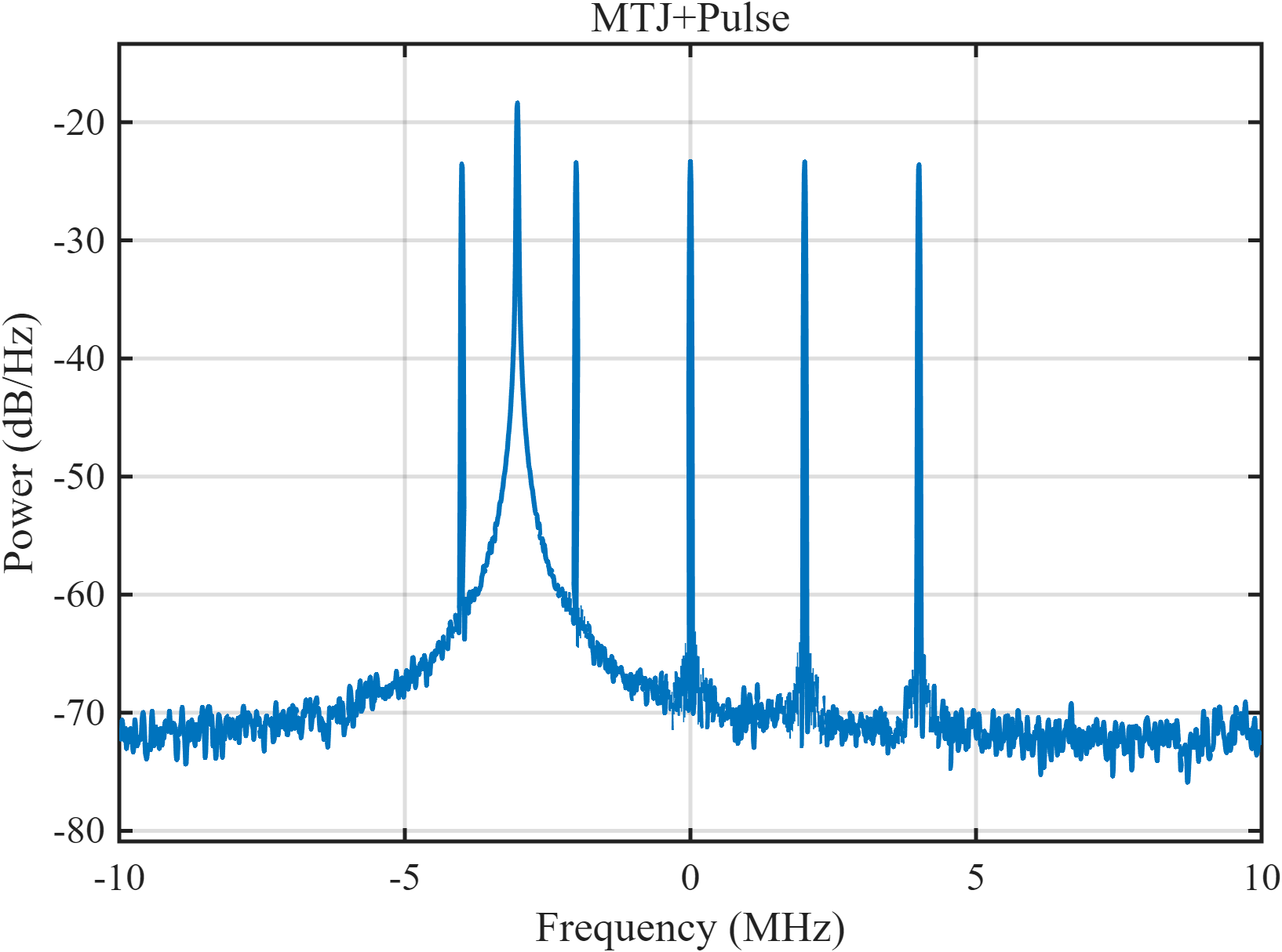}
\label{fig:psd_mtj_pulse}} \hfil
\subfloat[MTJ+PBNJ]{\includegraphics[width=0.29\linewidth]{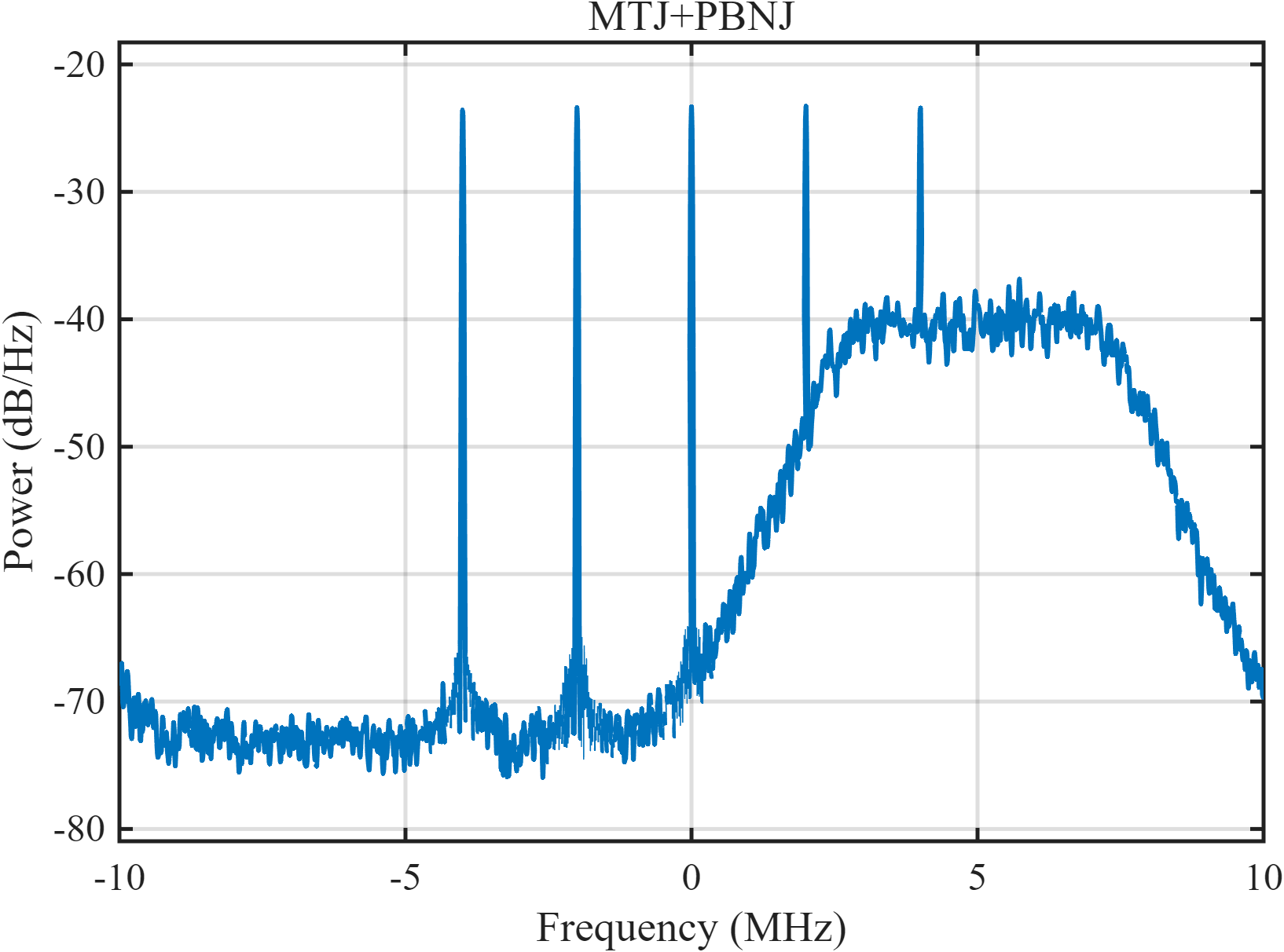}
\label{fig:psd_mtj_pbnj}} \\

\vspace{1em}
\subfloat[LFM+Pulse]{\includegraphics[width=0.29\linewidth]{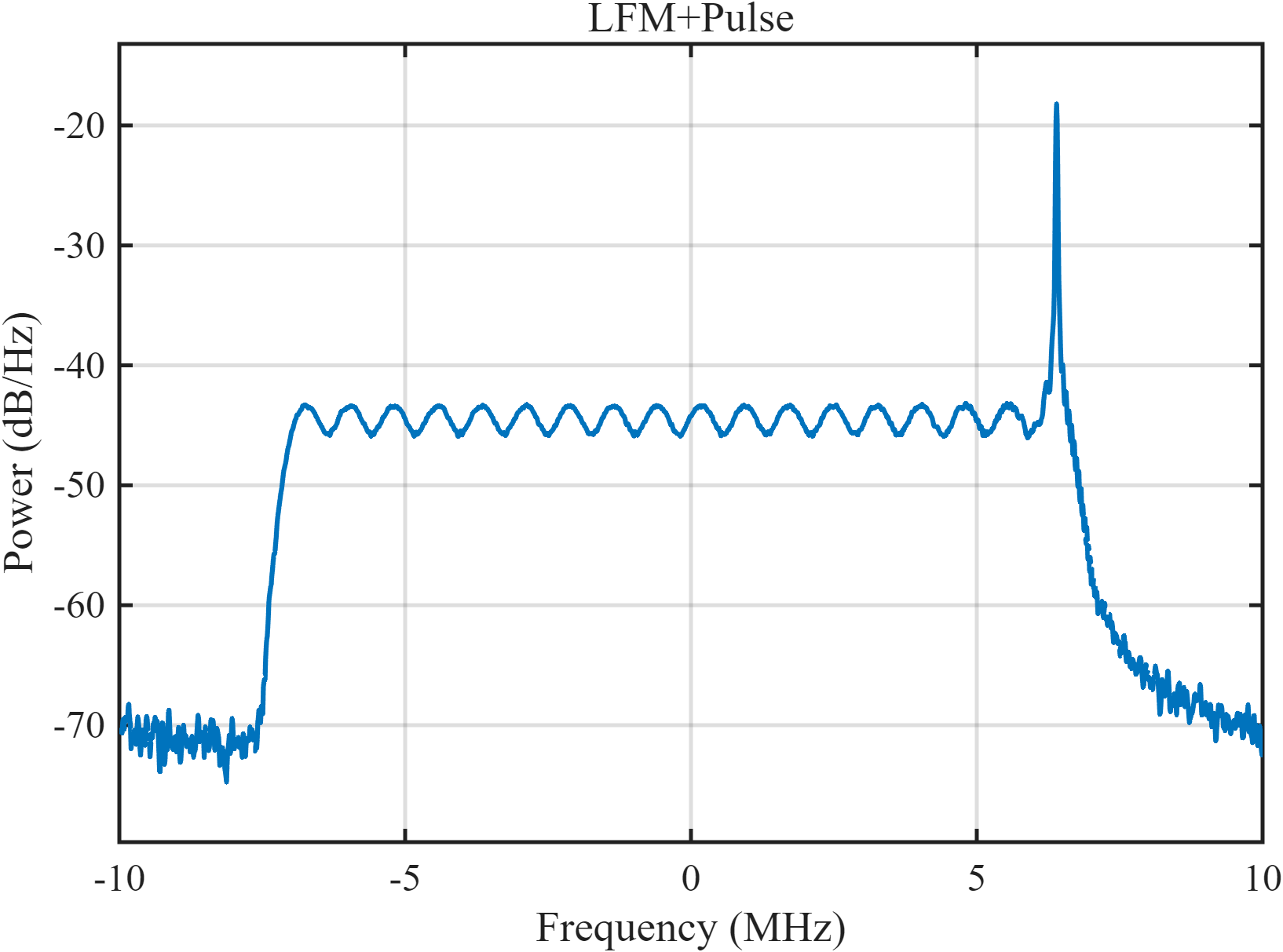}
\label{fig:psd_lfm_pulse}} \hfil
\subfloat[LFM+PBNJ]{\includegraphics[width=0.29\linewidth]{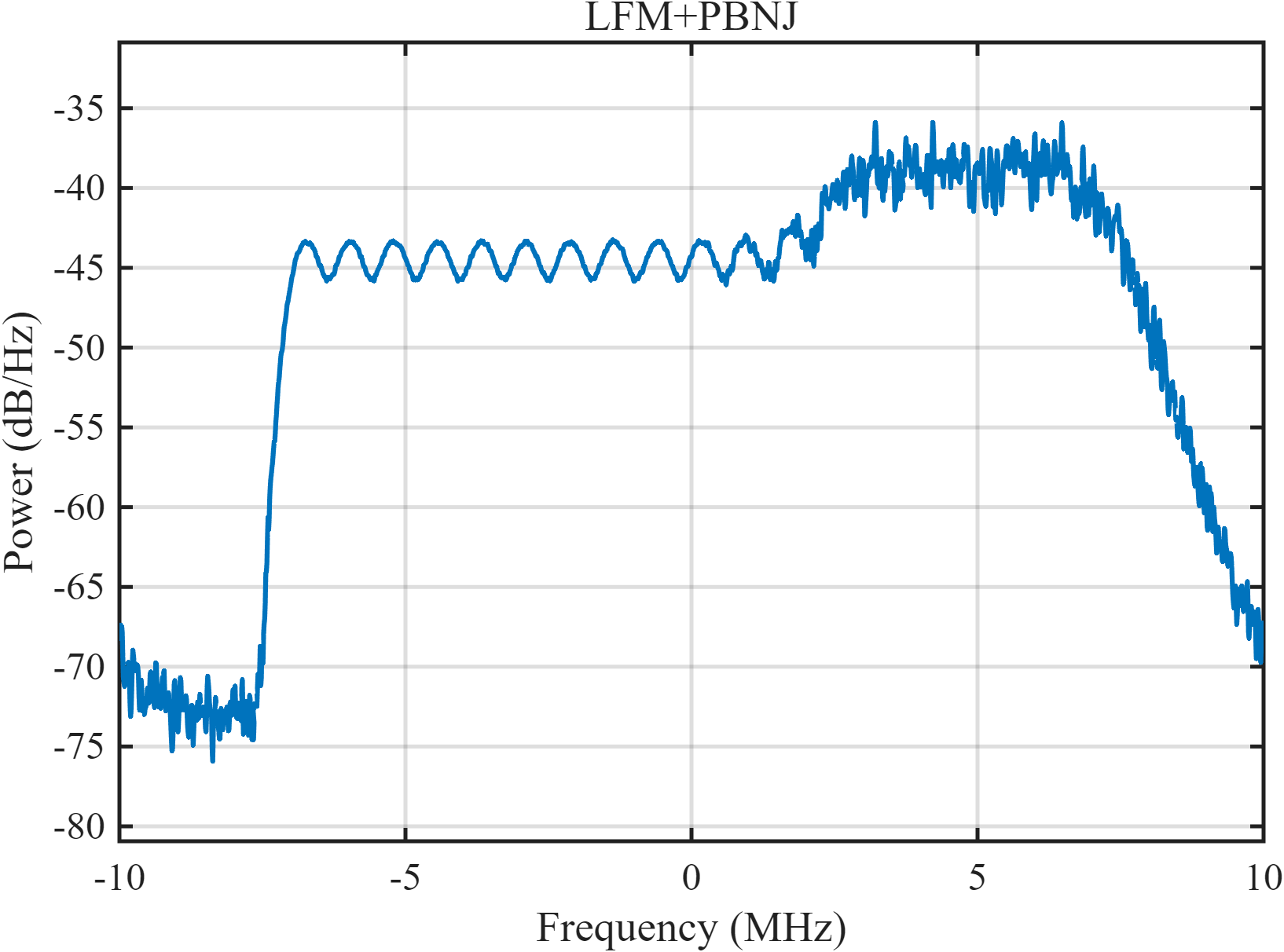}
\label{fig:psd_lfm_pbnj}} \hfil
\subfloat[Pulse+PBNJ]{\includegraphics[width=0.29\linewidth]{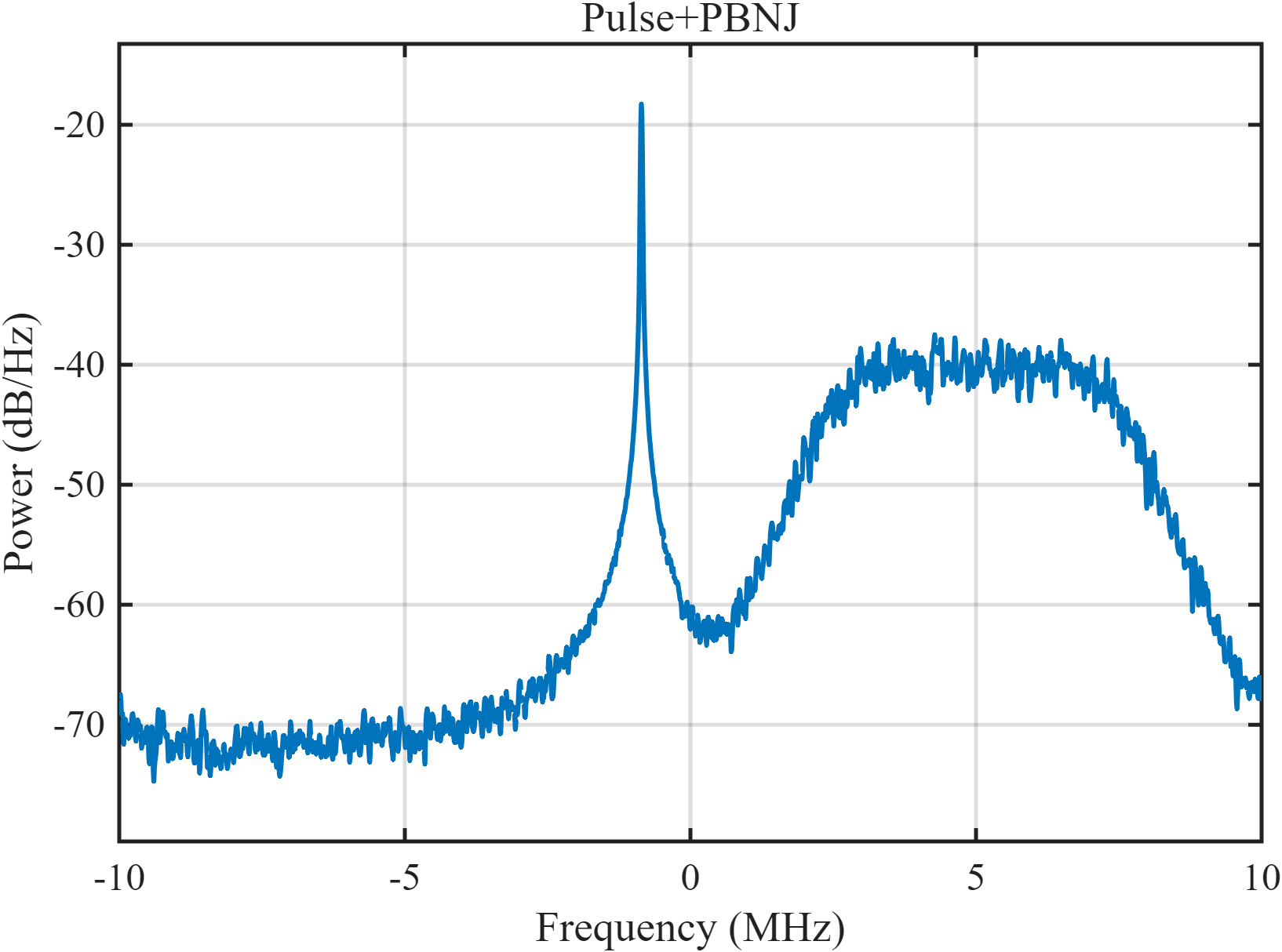}
\label{fig:psd_pulse_pbnj}}

\caption{Power Spectral Density (PSD) analysis of the 9 compound interference types. The frequency domain characteristics provide complementary energy distribution profiles, facilitating in the discrimination of spectrally similar compound signals.}
\label{fig:compound_psd}
\end{figure}


\section{Methodology}
\label{sec:methodology}

To address the challenge of classifying compound GNSS jamming signals under low JNR conditions, we propose SKANet. This architecture combines time-frequency features (STFT) and frequency-domain energy distributions (PSD) through an adaptive attention mechanism.

\subsection{Problem Formulation}
\label{subsec:problem_formulation}

We formulate the identification of compound GNSS jamming signals as a supervised classification problem under a closed-set assumption. In this context, the receiver is tasked with categorizing incoming signals against a pre-defined library of jamming primitives and their combinations, as established in the taxonomy presented in Section \ref{sec:system_model}. Unlike unsupervised clustering approaches that attempt to discover the number of signal clusters from unlabelled data, our framework leverages expert domain knowledge regarding jamming physics. Specifically, the total number of interference classes, denoted as $K$, is determined by the specific jamming combinations considered in the system model (i.e., $K=9$ in this study, corresponding to the types listed in Table \ref{tab:compound_params_full}).

The framework processes the signal into two complementary representations: the TFI, denoted as $\mathbf{X}_{\text{TFI}} \in \mathbb{R}^{H \times W \times C}$, and the PSD map, denoted as $\mathbf{X}_{\text{PSD}} \in \mathbb{R}^{H \times W \times C}$. Accordingly, the dataset is defined as $\mathcal{D} = \{(\mathbf{X}_{\text{TFI}}^{(i)}, \mathbf{X}_{\text{PSD}}^{(i)}, y^{(i)})\}_{i=1}^{N}$, where $N$ represents the total number of samples. The scalar $y^{(i)} \in \{1, 2, \dots, K\}$ denotes the ground-truth label associated with the $i$-th sample.

The core objective is to approximate the optimal mapping function $\mathcal{F}: (\mathbf{X}_{\text{TFI}}, \mathbf{X}_{\text{PSD}}) \to \hat{\mathbf{y}}$ that accurately predicts the probability distribution of the jamming categories within the known threat space. Reflecting the architecture of our proposed SKANet, the predictive model is mathematically decomposed as:
\begin{equation}
    \hat{\mathbf{y}} = \mathcal{F}(\mathbf{X}_{\text{TFI}}, \mathbf{X}_{\text{PSD}}; \Theta) = \mathcal{G}_{\text{cls}} \left( \mathcal{H}_{\text{fuse}} \left( \Phi_{\text{S}}(\mathbf{X}_{\text{TFI}}), \Phi_{\text{P}}(\mathbf{X}_{\text{PSD}}) \right) \right),
    \label{eq:model_structure}
\end{equation}

\begin{figure*}[!t]
    \centering
    \includegraphics[width=0.9\linewidth]{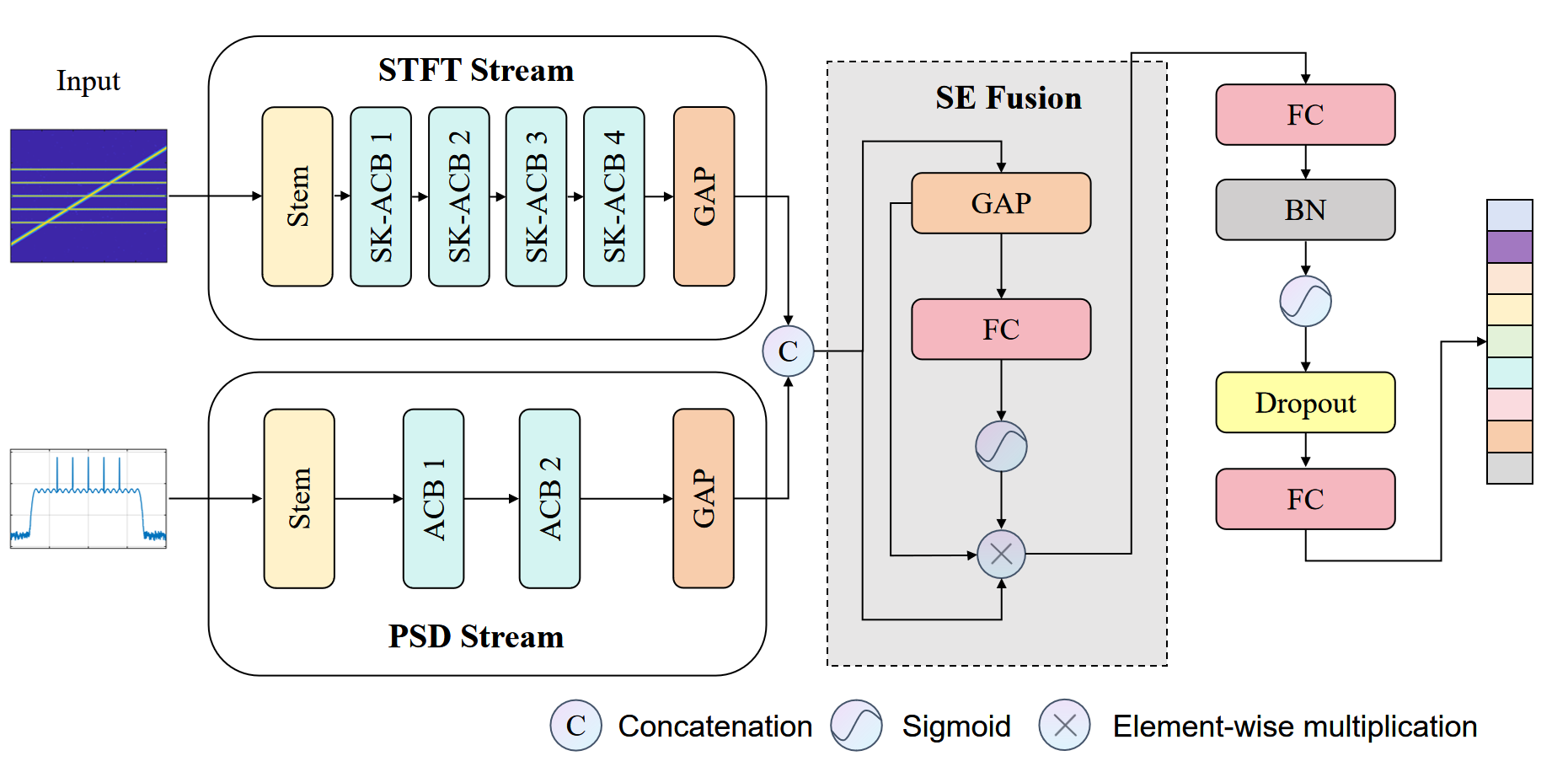}
    \caption{Schematic of the proposed SKANet architecture. The framework features a cognitive dual-stream design to process heterogeneous features. The STFT Stream utilizes a deep hierarchical backbone with Multi-Branch SK-ACB modules to extract high-level semantic representations from Time-Frequency Images. The PSD Stream employs a lightweight shallow network to capture global energy statistics from Power Spectral Density. A Squeeze-and-Excitation (SE) fusion block adaptively recalibrates and integrates the multi-modal features via concatenation and element-wise multiplication before the final classification head.}
    \label{fig:network}
\end{figure*}

where $\Phi_{\text{S}}(\cdot)$ and $\Phi_{\text{P}}(\cdot)$ represent the non-linear feature extraction transformations parameterized by the STFT-stream and PSD-stream weights, respectively. $\mathcal{H}_{\text{fuse}}(\cdot)$ denotes the fusion mechanism (e.g., Squeeze-and-Excitation) that integrates the heterogeneous features, and $\mathcal{G}_{\text{cls}}(\cdot)$ serves as the final classifier mapping the latent representation to the probability space. The set $\Theta$ encompasses all learnable parameters within the network.

To optimize $\Theta$, we employ the Cross-Entropy (CE) loss function to quantify the divergence between the predicted probability distribution $\hat{\mathbf{y}}$ and the true label $y$. The empirical risk minimization objective is formulated as:
\begin{equation}
    \mathcal{L}(\Theta) = -\frac{1}{N} \sum_{i=1}^{N} \sum_{k=1}^{K} \mathbb{I}(y^{(i)} = k) \log \left( [\hat{\mathbf{y}}^{(i)}]_k \right),
    \label{eq:ce_loss}
\end{equation}
where $\mathbb{I}(\cdot)$ is the indicator function, and $[\hat{\mathbf{y}}^{(i)}]_k$ is the predicted probability for class $k$.

Given the high-dimensional and non-convex nature of the loss landscape in deep neural networks, we utilize the Adaptive Moment Estimation (Adam) \cite{ref39} algorithm for optimization. Adam accelerates convergence by adapting the learning rate for each parameter based on the first and second moments of the gradients. Let $g_t = \nabla_{\Theta} \mathcal{L}(\Theta_{t-1})$ be the gradient at time step $t$. The algorithm maintains exponential moving averages of the gradients ($m_t$) and the squared gradients ($v_t$) as follows:
\begin{align}
    m_t &= \beta_1 m_{t-1} + (1 - \beta_1) g_t, \label{eq:adam_m} \\
    v_t &= \beta_2 v_{t-1} + (1 - \beta_2) g_t^2, \label{eq:adam_v}
\end{align}
where $\beta_1$ and $\beta_2$ are the decay rates for the moment estimates.
The parameters are subsequently updated using the bias-corrected estimates $\hat{m}_t$ and $\hat{v}_t$ as follows:
\begin{equation}
    \Theta_t = \Theta_{t-1} - \eta \frac{\hat{m}_t}{\sqrt{\hat{v}_t} + \epsilon},
    \label{eq:adam_update}
\end{equation}
where $\eta$ is the learning rate and $\epsilon$ is a smoothing term to ensure numerical stability. The specific hyperparameter settings for the optimizer are detailed in Section \ref{sec:results}.

\subsection{SKANet Architecture}

To robustly classify compound jamming signals characterized by time-varying non-stationarity and multi-scale spectral features, we propose the Dual-Stream SK-ACB-Net. This architecture utilizes an asymmetric design, tailored for the distinct feature characteristics of TF spectrograms and PSD inputs. As depicted in Fig. \ref{fig:network}, the framework integrates three distinct components: a Multi-Branch SK-ACB module for dynamic receptive field adaptation, a deepened asymmetric backbone for hierarchical feature extraction, and an SE fusion mechanism for adaptive inter-modality recalibration.

\subsubsection{Multi-Branch SK-ACB Module Formulation}
The fundamental building block of our network is the SK-ACB module. This module combines the rotational invariance and skeletal robustness of ACBs with the dynamic receptive field selection mechanism of SK networks.

\paragraph{Asymmetric Convolution Formulation}
Standard square convolutions often exhibit redundancy in the kernel corners. To enhance the feature extraction capability of the central skeleton without increasing inference latency, we replace standard convolutions with ACBs. For a given input tensor $\mathbf{X}$, the ACB approximates a standard $3\times3$ convolution by aggregating the outputs of three parallel kernels. The output is formulated as:
\begin{align}
    \mathbf{Y}_{ACB} = \sigma\Big(&\mathcal{BN}(\mathbf{K}_{3\times3} * \mathbf{X}) + \mathcal{BN}(\mathbf{K}_{1\times3} * \mathbf{X}) \nonumber\\
    &+ \mathcal{BN}(\mathbf{K}_{3\times1} * \mathbf{X})\Big),
\end{align}
where $\mathcal{BN}$ represents Batch Normalization, and $\sigma(\cdot)$ denotes the Swish activation function \cite{ref40}, defined as $\sigma(x) = x \cdot \text{sigmoid}(x)$. Exploiting the additivity of convolution, these kernels are fused into a single equivalent kernel $\tilde{\mathbf{K}}$ during the inference phase to ensure zero computational overhead:
\begin{equation}
    \tilde{\mathbf{K}} = \mathbf{K}_{3\times3} \oplus \mathbf{K}_{1\times3} \oplus \mathbf{K}_{3\times1},
    \label{eq:kernel_fusion}
\end{equation}
where the operator $\oplus$ denotes the element-wise summation of the effective kernel weights. Specifically, it implies that the asymmetric kernels ($\mathbf{K}_{1\times3}$ and $\mathbf{K}_{3\times1}$) are first fused with their corresponding batch normalization parameters and then zero-padded to the $3\times3$ spatial dimension to align with $\mathbf{K}_{3\times3}$ before aggregation.

Standard $3 \times 3$ convolutions are isotropic. However, jamming signals exhibit strong directionality in the Time-Frequency domain. For instance, STJ manifests as continuous horizontal lines characterized by temporal consistency, which are effectively captured by $1 \times 3$ kernels. Conversely, pulse jamming and wideband bursts appear as vertical structures representing spectral spread, which resonate with $3 \times 1$ kernels. The ACB thus acts as a physically-matched filter bank.

\begin{figure*}[!t]
    \centering
        \centering
        \includegraphics[width=0.90\linewidth]{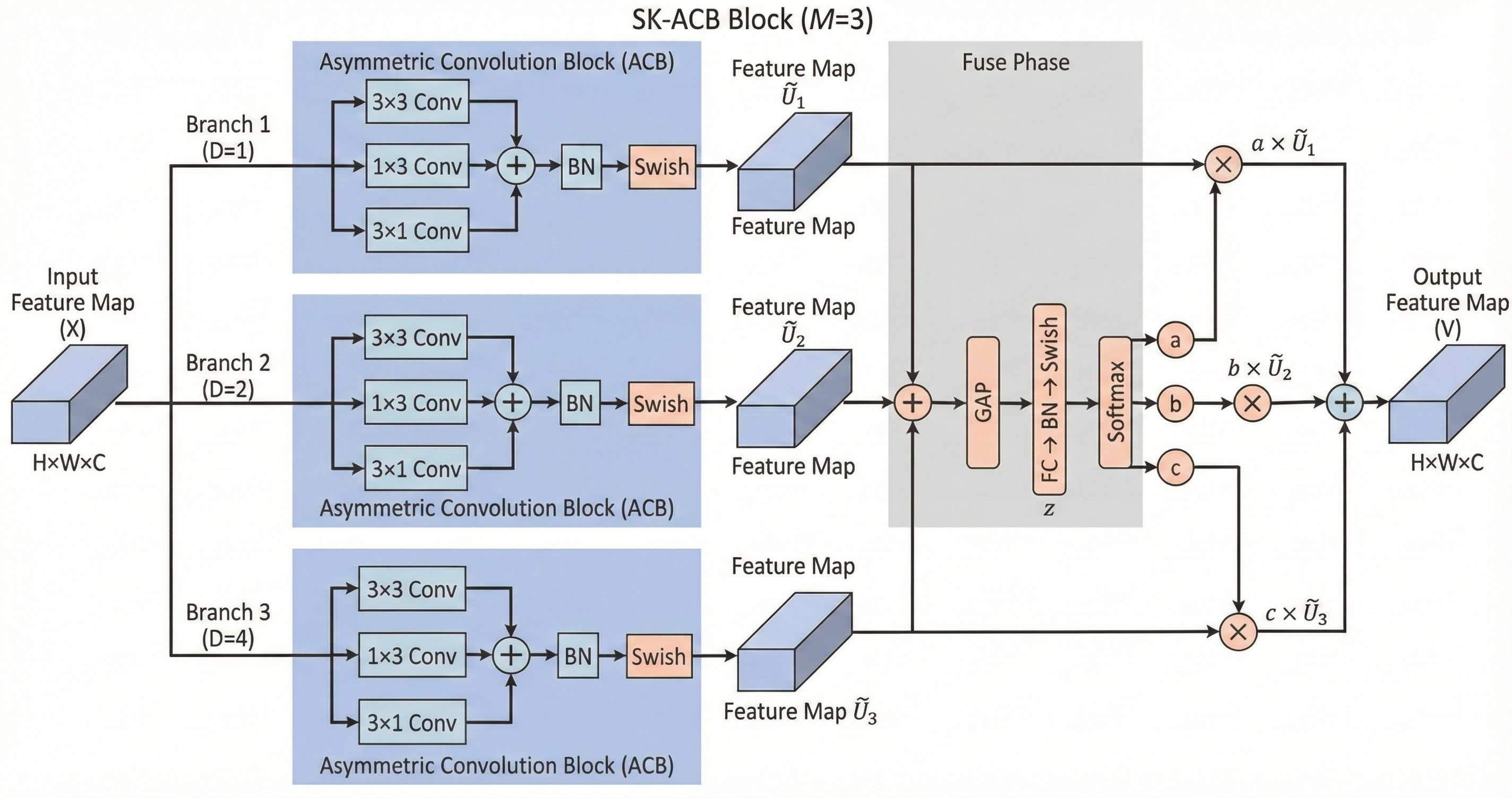}
    \caption{Schematic of the proposed method. The detailed internal structure of the SK-ACB block. It employs three parallel branches with distinct dilation rates ($d=1, 2, 4$) and Asymmetric Convolution Blocks (ACBs) to adaptively aggregate multi-scale features via a Split-Fuse-Select mechanism.}
    \label{fig:network_arch}
\end{figure*}

\paragraph{Adaptive Receptive Field Selection}
To simultaneously capture narrow-band impulses (e.g., pulse jamming) and wide-band spectral sweeps (e.g., LFM), the network requires adaptable receptive fields. We implement a Split-Fuse-Select mechanism with $M=3$ branches, as illustrated in Fig. \ref{fig:network_arch}. In the \textit{Split} phase, the input $\mathbf{X}$ is transformed into three feature maps $\tilde{\mathbf{U}}_m$ via parallel ACBs with distinct dilation rates $D=\{1, 2, 4\}$, constructing a pyramid of receptive fields that covers local textures to global contexts. These features are then \textit{Fused} via element-wise summation and compressed into a channel-wise statistic vector $\mathbf{s}$ using Global Average Pooling (GAP). Finally, in the \textit{Select} phase, a compact descriptor $\mathbf{z}$ is generated via a Fully Connected (FC) layer to compute adaptive attention weights $\mathbf{a}$, $\mathbf{b}$ and $\mathbf{c}$ through a Softmax operator. The final output $\mathbf{V}$ is obtained by the weighted aggregation of the multi-scale features, mathematically expressed as follows:
\begin{align}
    & \mathbf{U} = \sum_{m=1}^{M} \tilde{\mathbf{U}}_m, \label{eq:sk_sum} \\
    & s_c = \frac{1}{H \times W} \sum_{i=1}^{H} \sum_{j=1}^{W} \mathbf{U}_c(i,j), \label{eq:sk_gap} \\
    & a_c = \frac{e^{\mathbf{A}_c \mathbf{z}}}{e^{\mathbf{A}_c \mathbf{z}} + e^{\mathbf{B}_c \mathbf{z}} + e^{\mathbf{C}_c \mathbf{z}}}, \label{eq:sk_softmax} \\
    & \mathbf{V}_c = a_c \cdot \tilde{\mathbf{U}}_{1,c} + b_c \cdot \tilde{\mathbf{U}}_{2,c} + c_c \cdot \tilde{\mathbf{U}}_{3,c}, \label{eq:sk_out}
\end{align}
where $\mathbf{A}$, $\mathbf{B}$ and $\mathbf{C}$ are learnable weight matrices, and the weights satisfy $a_c + b_c + c_c = 1$.

\subsubsection{Hierarchical Asymmetric Backbone}
Recognizing that TF spectrograms contain rich, non-stationary semantic information while PSD plots provide concise statistical summaries, we construct an asymmetric dual-stream backbone. The \textit{STFT Stream} serves as the primary deep feature extractor, initiating with a stem layer followed by four cascaded stages of SK-ACB blocks ($L_{stft}=4$). We employ a progressive channel expansion strategy ($32 \to 512$) to abstract high-level semantic representations. In contrast, the \textit{PSD Stream} is designed for computational efficiency; it utilizes a shallower architecture with only two processing stages and a channel capacity capped at 128. This design prevents overfitting to the simpler frequency-domain features while minimizing parameter redundancy.

\subsubsection{Squeeze-and-Excitation Fusion and Classification}
To synthesize the heterogeneous features, we employ a concatenation-based fusion augmented by an SE mechanism. The feature vectors $\mathbf{F}_{STFT} \in \mathbb{R}^{512}$ and $\mathbf{F}_{PSD} \in \mathbb{R}^{128}$ obtained after GAP are first concatenated into a unified representation $\mathbf{F}_{cat}$. An SE block is then applied to adaptively recalibrate channel importance by learning a weight vector $\mathbf{w}$ through a bottleneck structure with a reduction ratio of $r=16$. The final fused representation $\mathbf{F}_{final}$ is generated via channel-wise multiplication and subsequently fed into the classifier head. The classifier comprises a linear projection to 256 latent dimensions, Batch Normalization, Swish activation, and a Dropout layer ($p = 0.6$) before the final probability mapping. This dropout rate was determined via hyperparameter tuning on the validation set. A rate of $0.6$ was empirically found effective to mitigate overfitting, specifically by preventing the network from memorizing noise patterns in low-JNR samples and reducing the feature redundancy inherent in the concatenation of dual-stream representations. The fusion and recalibration process is formulated as:
\begin{align}
    & \mathbf{F}_{cat} = \text{Concat}(\mathbf{F}_{STFT}, \mathbf{F}_{PSD}), \label{eq:fusion_cat} \\
    & \mathbf{w} = \sigma_{sig}\left(\mathbf{W}_2 \cdot \delta_{relu}(\mathbf{W}_1 \cdot \mathbf{F}_{cat})\right), \label{eq:fusion_w} \\
    & \mathbf{F}_{final} = \mathbf{w} \odot \mathbf{F}_{cat}, \label{eq:fusion_final}
\end{align}
where $\delta_{relu}$ is the ReLU activation, $\sigma_{sig}$ is the Sigmoid function, $\mathbf{W}_1, \mathbf{W}_2$ represent the weights of the fully connected layers in the SE block, and the operator $\odot$ denotes the element-wise multiplication (Hadamard product).


\section{Simulation Results and Analysis}
\label{sec:results}

In this section, we present an evaluation of the proposed SKANet. We first detail the dataset generation and experimental setup, followed by an analysis of the classification performance.

\subsection{Dataset Generation and Preprocessing}
To evaluate the proposed method, we constructed a synthetic dataset simulating compound GNSS interference scenarios. The simulation environment is configured to replicate the signal characteristics of the GPS L1 frequency band ($1575.42$ MHz), ensuring the spectral behaviors of the generated jamming signals are physically representative\cite{ref27} .

\subsubsection{Signal Simulation Parameters}
The baseband simulation is conducted with a sampling frequency of $F_s = 20$ MHz, covering the typical bandwidth of civil GNSS receivers\cite{ref6, ref18}. The observation duration is set to $T = 1$ ms \cite{ref19, ref25}, corresponding to one C/A code period, yielding a discrete sequence length of $N = 20,000$ points. To evaluate the robustness of the model under varying jamming intensities, the JNR is systematically swept from $-25$ dB to $15$ dB in $1$ dB increments. The background noise is modeled as complex AWGN\cite{ref18, ref41}.

\subsubsection{Compound Interference Configuration}
This study focuses on ``dual-component'' compound interference. Following the superposition mechanism defined in \eqref{eq:compound}, compound jamming signals are synthesized by linearly combining a primary and a secondary jamming primitive. To simulate realistic power imbalances encountered in dynamic electromagnetic environments, the PR, as formulated in \eqref{eq:PR}, is randomized uniformly within the range of $[-3, 3]$ dB for each sample.

We consider a total of nine distinct compound interference classes formed by the pairwise superposition of jamming primitives. These classes encompass interactions between narrow-band, wide-band, and sweep interferences, specifically: STJ+LFM, STJ+Pulse, STJ+PBNJ, MTJ+LFM, MTJ+Pulse, MTJ+PBNJ, LFM+Pulse, LFM+PBNJ, and Pulse+PBNJ.

\subsubsection{Dual-Domain Feature Generation}
To feed the dual-stream architecture, each raw time-domain signal is transformed into two complementary visual representations:

\paragraph{Time-Frequency Image (TFI)}
The TFI is generated via STFT. To balance time and frequency resolution, we employ a Hann window. The window length $N_{win}$ is adaptively set to 128 for continuous signals and 64 for short-burst signals to capture transient details. The overlap is configured at approximately $92\%$, and a 4096-point FFT is applied. The resulting magnitude spectrogram is logarithmically scaled to dB and resized to $224 \times 224$ pixels using bicubic interpolation.

\paragraph{PSD Image}
The frequency-domain input is generated using Welch's power spectral density estimate. We utilize a Hamming window of length 2048 with $50\%$ overlap and a 4096-point FFT. This configuration provides a smooth and low-variance spectral estimate. The resulting PSD curve is plotted without axes or grid lines and captured as a $224 \times 224$ pixel image, strictly containing the spectral trace to minimize visual noise.

\subsubsection{Dataset Scale}
The final dataset comprises 405,000 samples. This includes nine compound jamming classes, with 1,000 independent Monte Carlo realizations generated for each class at each of the JNR levels ($-25$ dB to $15$ dB).

\begin{table}[!t]
\renewcommand{\arraystretch}{1.2}
\caption{Parameter Settings for The Compound Jamming Signals}
\label{tab:compound_params_full}
\centering
\resizebox{\linewidth}{!}{
\begin{tabular}{c c c}
\toprule
\textbf{Signals} & \textbf{Parameters} & \textbf{Values} \\
\midrule

\multirow{4}{*}{STJ+LFM}
 & Carrier Freq. of STJ & $\pm 9.5$ MHz \\
 & Initial Phase of STJ & $0 \sim 2\pi$ \\
 & Sweep Bandwidth of LFM & 10 MHz \\
 & Sweep Period of LFM & 1 ms \\
\midrule

\multirow{4}{*}{STJ+Pulse}
 & Carrier Freq. of STJ & $\pm 9.5$ MHz \\
 & Initial Phase of STJ & $0 \sim 2\pi$ \\
 & Duty Cycle of Pulse & $\approx 30\%$ \\
 & PRI of Pulse & $N/6$ samples \\
\midrule

\multirow{3}{*}{STJ+PBNJ}
 & Carrier Freq. of STJ & $\pm 9.5$ MHz \\
 & Initial Phase of STJ & $0 \sim 2\pi$ \\
 & Bandwidth of PBNJ & $10\% \sim 25\% F_s$ \\
\midrule

\multirow{4}{*}{MTJ+LFM}
 & Tone Number of MTJ & $3 \sim 6$ \\
 & Tone Spacing of MTJ & $1.5 \sim 3.0$ MHz \\
 & Sweep Bandwidth of LFM & 10 MHz \\
 & Sweep Period of LFM & 1 ms \\
\midrule

\multirow{4}{*}{MTJ+Pulse}
 & Tone Number of MTJ & $3 \sim 6$ \\
 & Tone Spacing of MTJ & $1.5 \sim 3.0$ MHz \\
 & Duty Cycle of Pulse & $\approx 30\%$ \\
 & PRI of Pulse & $N/6$ samples \\
\midrule

\multirow{3}{*}{MTJ+PBNJ}
 & Tone Number of MTJ & $3 \sim 6$ \\
 & Tone Spacing of MTJ & $1.5 \sim 3.0$ MHz \\
 & Bandwidth of PBNJ & $10\% \sim 25\% F_s$ \\
\midrule

\multirow{4}{*}{LFM+Pulse}
 & Sweep Bandwidth of LFM & 10 MHz \\
 & Sweep Period of LFM & 1 ms \\
 & Duty Cycle of Pulse & $\approx 30\%$ \\
 & PRI of Pulse & $N/6$ samples \\
\midrule

\multirow{3}{*}{LFM+PBNJ}
 & Sweep Bandwidth of LFM & 10 MHz \\
 & Sweep Period of LFM & 1 ms \\
 & Bandwidth of PBNJ & $10\% \sim 25\% F_s$ \\
\midrule

\multirow{3}{*}{Pulse+PBNJ}
 & Duty Cycle of Pulse & $\approx 30\%$ \\
 & PRI of Pulse & $N/6$ samples \\
 & Bandwidth of PBNJ & $10\% \sim 25\% F_s$ \\

\bottomrule
\end{tabular}
}
\end{table}

\subsection{Implementation Details}
\label{subsec:implementation}

\subsubsection{Hardware and Software Environment}
Experiments utilized a workstation equipped with an Intel Core i9-12900KF CPU (@ 3.20 GHz) and an NVIDIA GeForce RTX 5060 Ti GPU (16 GB VRAM). The DL models were implemented using the PyTorch 2.8.0 library within a Python 3.11 environment. All training and inference processes were accelerated using CUDA 11.8 and cuDNN libraries.

\subsubsection{Training Strategy}
The network was optimized using Stochastic Gradient Descent (SGD) with the Adam optimizer ($\beta_1=0.9$, $\beta_2=0.999$, and $\epsilon=10^{-8}$). To facilitate convergence, the learning rate was initialized at $\eta = 10^{-3}$ and adjusted dynamically via a cosine annealing scheduler over 100 training epochs. The model was trained with a batch size of 64, using CE loss to guide gradient updates. To mitigate overfitting, a dropout rate of $p=0.6$ was applied to the classifier head. The dataset was partitioned into training, validation, and testing sets following an 0.8:0.1:0.1 ratio. All reported results are averaged over 10 independent Monte Carlo runs.

\subsubsection{Signal Parameterization}
The synthetic dataset relies on the distinct physical characteristics of five fundamental jamming primitives. The specific parameters and value ranges for each interference type are detailed in Table \ref{tab:compound_params_full}. These parameters were randomized for each sample to ensure dataset diversity.

\subsection{Performance Evaluation Metrics}
\label{subsec:evaluation_metrics}

To assess the classification performance and resource utilization of the proposed model, we employ standard quantitative metrics: Overall Accuracy (OA), class-wise Precision and Recall, F1-Score, Confusion Matrix, and Floating Point Operations (FLOPs).

\subsubsection{Accuracy Metrics}
OA serves as the primary metric, quantifying the ratio of correctly classified samples to the total number of samples. Let $N$ be the total number of test samples, and $N_{correct}$ be the number of correct predictions. The OA is defined as:
\begin{equation}
    OA = \frac{N_{correct}}{N} \times 100\%.
\end{equation}

\subsubsection{Precision and Recall}
To evaluate the model's reliability on specific jamming classes, we utilize Precision and Recall. Precision ($P_k$) measures the accuracy of positive predictions for class $k$, while Recall ($R_k$) measures the ability to capture all positive instances. These are calculated based on True Positives ($TP_k$), False Positives ($FP_k$), and False Negatives ($FN_k$):
\begin{equation}
    P_k = \frac{TP_k}{TP_k + FP_k},
\end{equation}
\begin{equation}
    R_k = \frac{TP_k}{TP_k + FN_k}.
\end{equation}

\subsubsection{F1-Score}
Given the potential for varying difficulty across jamming types, the F1-Score provides a balanced metric by calculating the harmonic mean of Precision and Recall. It is particularly useful for assessing robustness on imbalanced classes:
\begin{equation}
    F1_k = 2 \cdot \frac{P_k \cdot R_k}{P_k + R_k}.
\end{equation}

\subsubsection{Confusion Matrix}
To visualize misclassification between jamming classes, we utilize the Confusion Matrix $\mathbf{M} \in \mathbb{R}^{K \times K}$, where $K=9$ is the number of classes. The element $M_{ij}$ represents the number of samples belonging to class $i$ that are predicted as class $j$. Diagonal elements $M_{ii}$ correspond to correct predictions, while off-diagonal elements indicate specific error patterns.

\subsubsection{Computational Complexity}
To evaluate computational efficiency and hardware resource utilization independent of specific platforms, we quantify complexity using the total number of Floating Point Operations (FLOPs). Given that the SKANet architecture includes deep convolutional backbones and dense classification heads, we adopt a rigorous calculation where each Multiply-Accumulate (MAC) operation is counted as two floating-point operations.

Let $L$ denote the total number of convolutional layers and $Q$ denote the number of linear (fully connected) layers. The complexity for the convolutional component, $\mathcal{F}_{conv}$ (in FLOPs), is calculated by summing the operations over all layers $i=1 ,\dots, L$:
\begin{equation}
    \mathcal{F}_{conv} = \sum_{i=1}^{L} 2 \cdot H_i W_i \cdot K_i^2 \cdot C_{in}^{(i)} \cdot C_{out}^{(i)},
\end{equation}
where $H_i$ and $W_i$ represent the height and width of the output feature map at the $i$-th layer, respectively. $K_i$ denotes the spatial size of the square convolution kernel, while $C_{in}^{(i)}$ and $C_{out}^{(i)}$ correspond to the number of input and output channels. The factor $2$ explicitly accounts for the one multiplication and one addition inherent in each MAC operation.

Similarly, the complexity for the linear layers, $\mathcal{F}_{lin}$, depends on the number of input neurons $N_{in}^{(j)}$ and output neurons $N_{out}^{(j)}$ for each layer $j = 1, \ldots, Q$. This is formulated as:
\begin{equation}
    \mathcal{F}_{lin} = \sum_{j=1}^{Q} (2 N_{in}^{(j)} - 1) \cdot N_{out}^{(j)}.
\end{equation}

This formulation captures the computational cost of vector dot products, where generating each of the $N_{out}^{(j)}$ output values requires $N_{in}^{(j)}$ multiplications and $N_{in}^{(j)} - 1$ additions. The total network complexity is the sum of these two components.

\subsection{Performance Analysis}
\label{subsec:performance_analysis}

To rigorously evaluate the efficacy of SKANet, we conduct a comparative analysis against a set of mainstream Deep Learning baselines. The benchmark models include the Transformer architecture ViT-B-16 and several Convolutional Neural Network architectures including ResNet18, EfficientNetB0, ShuffleNetV2, and the domain-specific TFPENet.

Regarding the experimental configuration, the statistical metrics reported, including Overall Accuracy and F1 Score, represent aggregate performance averaged over the JNR range of $-20$ dB to $15$ dB. We explicitly exclude the extreme scenario of $-25$ dB from the aggregate metrics. At a Jamming-to-Noise Ratio of $-25$ dB, the energy of the jamming signal is deeply submerged below the thermal noise floor which renders intrinsic spectral features virtually indistinguishable. Under such extreme conditions, the recognition capability of all evaluated classifiers inevitably degrades toward random guessing and lacks statistical significance for comparative analysis.

\begin{table}[!t]
\centering
\caption{Comparison of Computational Complexity and Overall Performance (Averaged over JNR $-20$ dB to $15$ dB)}
\label{tab:complexity_comparison}
\renewcommand{\arraystretch}{1.3}
\resizebox{0.9\columnwidth}{!}{%
\begin{tabular}{cccc}
\toprule
Methods & OA (\%) & Params (M) & Time (ms) \\ \midrule
ViT-B-16 & 86.92 & 85.81 & 6.65 \\
RCNN & 88.64 & 13.12 & 0.39 \\
ShuffleNetV2 \cite{ref37}& 90.88 & 1.26 & 3.67 \\
ResNet18 \cite{ref34}& 91.54 & 11.18 & 1.36 \\
EfficientNetB0 \cite{ref35}& 92.74 & 4.02 & 5.00 \\
TFPENet \cite{ref10}& 94.58 & 11.57 & 4.47 \\
\textbf{SKANet} & \textbf{96.99} & \textbf{10.63} & \textbf{5.38} \\ \bottomrule
\end{tabular}%
}
\end{table}

\begin{table*}[!t]
\centering
\caption{Precision(\%), Recall(\%), and F1 Score(\%) of Various Recognition Methods (Averaged over JNR $-20$ dB to $15$ dB)}
\label{tab:metrics}
\renewcommand{\arraystretch}{1.3}
\resizebox{\textwidth}{!}{%
\begin{tabular}{c|ccc|ccc|ccc|ccc|ccc}
\toprule
\multirow{2}{*}{Methods} & \multicolumn{3}{c|}{LFM\_PBNJ} & \multicolumn{3}{c|}{LFM\_Pulse} & \multicolumn{3}{c|}{MTJ\_LFM} & \multicolumn{3}{c|}{MTJ\_PBNJ} & \multicolumn{3}{c}{MTJ\_Pulse} \\ \cline{2-16}
 & Pr & R & F1 & Pr & R & F1 & Pr & R & F1 & Pr & R & F1 & Pr & R & F1 \\ \midrule
RCNN & 87.25 & 86.15 & 86.68 & 88.04 & 89.05 & 88.52 & 86.82 & 87.20 & 86.98 & 89.15 & 88.60 & 88.85 & 87.66 & 88.85 & 88.22 \\
ShuffleNetV2 & 88.50 & 81.35 & 84.77 & 91.15 & 90.10 & 90.62 & 85.90 & 88.55 & 87.20 & 87.45 & 88.90 & 88.17 & 89.80 & 90.45 & 90.12 \\
ResNet18 & 89.15 & 89.20 & 89.17 & 92.50 & 92.15 & 92.32 & 88.60 & 88.45 & 88.52 & 87.20 & 86.80 & 87.00 & 90.10 & 91.65 & 90.87 \\
ViT-B-16 & 88.35 & 79.30 & 83.58 & 87.10 & 90.15 & 88.60 & 83.25 & 88.50 & 85.79 & 85.40 & 85.20 & 85.30 & 84.60 & 87.85 & 86.19 \\
EfficientNetB0 & 91.36 & 89.65 & 90.50 & 93.85 & 93.06 & 93.45 & 90.90 & 90.85 & 90.87 & 92.20 & 89.58 & 90.87 & 90.50 & 93.86 & 92.15 \\
TFPENet & \underline{97.45} & \textbf{95.10} & \textbf{96.26} & \underline{96.20} & \underline{94.65} & \underline{95.42} & \underline{93.85} & \underline{93.20} & \underline{93.52} & \underline{96.10} & \underline{93.95} & \underline{95.01} & \underline{96.55} & \underline{97.35} & \underline{96.95} \\
\textbf{SKANet} & \textbf{99.37} & \underline{94.36} & \underline{96.20} & \textbf{96.46} & \textbf{95.87} & \textbf{96.15} & \textbf{94.52} & \textbf{96.26} & \textbf{94.64} & \textbf{96.88} & \textbf{95.06} & \textbf{95.54} & \textbf{96.68} & \textbf{99.09} & \textbf{97.76} \\ \bottomrule
\toprule
\multirow{2}{*}{Methods} & \multicolumn{3}{c|}{Pulse\_PBNJ} & \multicolumn{3}{c|}{STJ\_LFM} & \multicolumn{3}{c|}{STJ\_PBNJ} & \multicolumn{3}{c|}{STJ\_Pulse} & \multicolumn{3}{c}{Average} \\ \cline{2-16}
 & Pr & R & F1 & Pr & R & F1 & Pr & R & F1 & Pr & R & F1 & Pr & R & F1 \\ \midrule
RCNN & 88.42 & 89.40 & 88.88 & 89.55 & 88.38 & 88.94 & 92.10 & 91.80 & 91.92 & 88.35 & 89.12 & 88.71 & 88.59 & 88.73 & 88.63 \\
ShuffleNetV2 & 88.65 & 90.65 & 89.64 & 90.20 & 89.15 & 89.67 & 92.45 & 92.80 & 92.62 & 88.10 & 92.35 & 90.17 & 89.13 & 89.37 & 89.22 \\
ResNet18 & 89.85 & 90.40 & 90.12 & 92.30 & 90.80 & 91.54 & 93.10 & 94.65 & 93.87 & 91.55 & 92.20 & 91.87 & 90.48 & 90.70 & 90.59 \\
ViT-B-16 & 85.15 & 81.10 & 83.08 & 86.40 & 88.65 & 87.51 & 87.25 & 87.40 & 87.32 & 89.10 & 89.20 & 89.15 & 86.29 & 86.37 & 86.28 \\
EfficientNetB0 & 90.95 & 90.91 & 90.93 & 94.24 & 91.95 & 93.08 & 93.73 & 94.55 & 94.14 & 94.01 & 95.85 & 94.92 & 92.42 & 92.25 & 92.32 \\
TFPENet & \underline{93.90} & \textbf{96.85} & \underline{95.35} & \underline{95.80} & \underline{94.60} & \underline{95.20} & \underline{97.10} & \underline{95.15} & \underline{96.11} & \underline{96.65} & \underline{96.40} & \underline{96.52} & \underline{95.96} & \underline{95.25} & \underline{95.59} \\
\textbf{SKANet} & \textbf{96.71} & \underline{94.91} & \textbf{95.72} & \textbf{96.26} & \textbf{97.55} & \textbf{96.88} & \textbf{97.20} & \textbf{95.81} & \textbf{96.46} & \textbf{98.25} & \textbf{99.82} & \textbf{98.98} & \textbf{96.73} & \textbf{96.53} & \textbf{96.48} \\ \bottomrule
\end{tabular}%
}
\end{table*}

\subsubsection{Overall Performance and Complexity Trade-off}
Table \ref{tab:complexity_comparison} presents a comprehensive evaluation of classification accuracy alongside computational overhead. SKANet achieves a superior Overall Accuracy of 96.99\% and establishes a lead of 2.41\% over the second-best method TFPENet. In terms of efficiency, while lightweight architectures like ShuffleNetV2 and ResNet18 offer lower parameter counts or reduced inference latency, their simplified feature extraction mechanisms result in compromised classification accuracy which is particularly evident in complex jamming scenarios. Conversely, the ViT-B-16 incurs the highest computational cost with over 85 million parameters yet fails to deliver commensurate performance gains.

SKANet demonstrates an optimal equilibrium between precision and efficiency. Compared to TFPENet, our proposed model reduces the parameter count by approximately 0.9 million while achieving a substantial accuracy improvement. Although the dual-stream attention mechanism introduces a marginal increase in inference time of approximately 0.9 ms per sample, this latency remains well within acceptable bounds for real-time CR applications and validates the practical viability of the proposed architecture.

\subsubsection{Robustness Across JNR Regimes}
\begin{figure}[!t]
    \centering
    \includegraphics[width=3.3in]{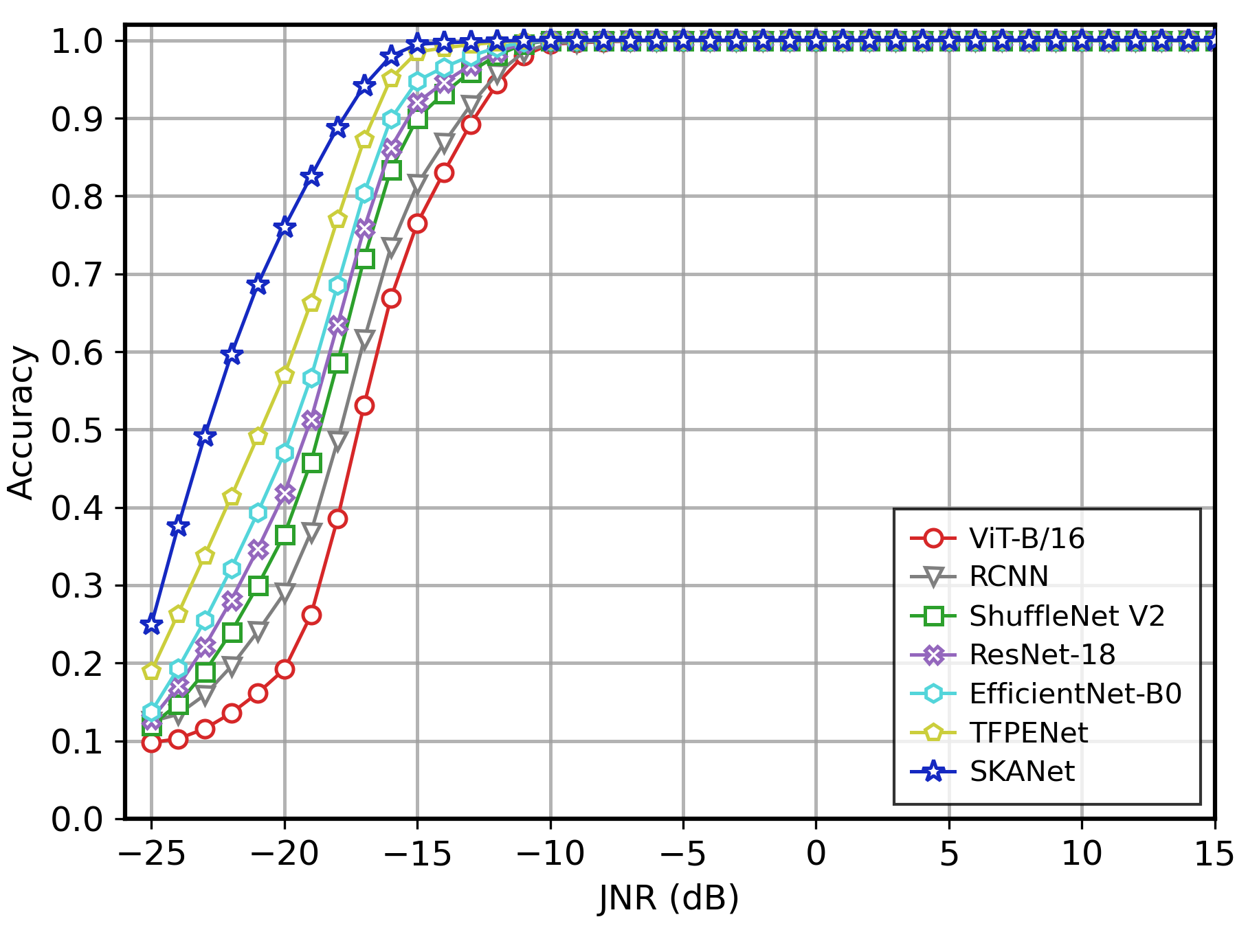}
    \caption{Classification accuracy comparison versus JNR.}
    \label{fig:accuracy}
\end{figure}

The evolution of classification accuracy with respect to varying Jamming-to-Noise Ratio levels is illustrated in Fig. \ref{fig:accuracy}. The curves reveal distinct performance characteristics across three signal quality regimes.

In the low-SNR regime ranging from $-25$ dB to $-15$ dB, SKANet exhibits exceptional robustness. At the challenging level of $-20$ dB, SKANet maintains a recognition accuracy exceeding 75\%. In contrast, standard convolutional baselines such as ResNet18 and ShuffleNetV2 struggle significantly with accuracies falling below 45\% and 40\% respectively. This substantial performance gap highlights the efficacy of the SK-ACB module in extracting salient features from noise-corrupted signals. The Transformer-based ViT-B-16 yields the lowest performance in this region. This limitation is attributable to the lack of inductive bias in standard Vision Transformers which renders them less effective at capturing local spectral-temporal primitives when the global structure is masked by heavy background noise.

In the transition regime between $-15$ dB and $-10$ dB, all models exhibit a rapid ascent in accuracy. However, SKANet demonstrates the fastest convergence rate and reaches near-perfect accuracy exceeding 99\% at $-14$ dB. TFPENet and EfficientNetB0 require higher signal power to achieve saturation and converge at approximately $-12$ dB and $-10$ dB respectively.

In the high-SNR regime greater than $-10$ dB, the performance of most models stabilizes. SKANet maintains a consistent 100\% accuracy which confirms that the model does not suffer from overfitting or degradation when signal conditions improve. These observations confirm that SKANet not only provides high accuracy across a wide dynamic range but also extends the operational limit of the receiver into lower Jamming-to-Noise Ratio environments.

\subsubsection{Inter-Class Discriminability Analysis}
\begin{figure}[!t]
    \centering
    \includegraphics[width=3.2in]{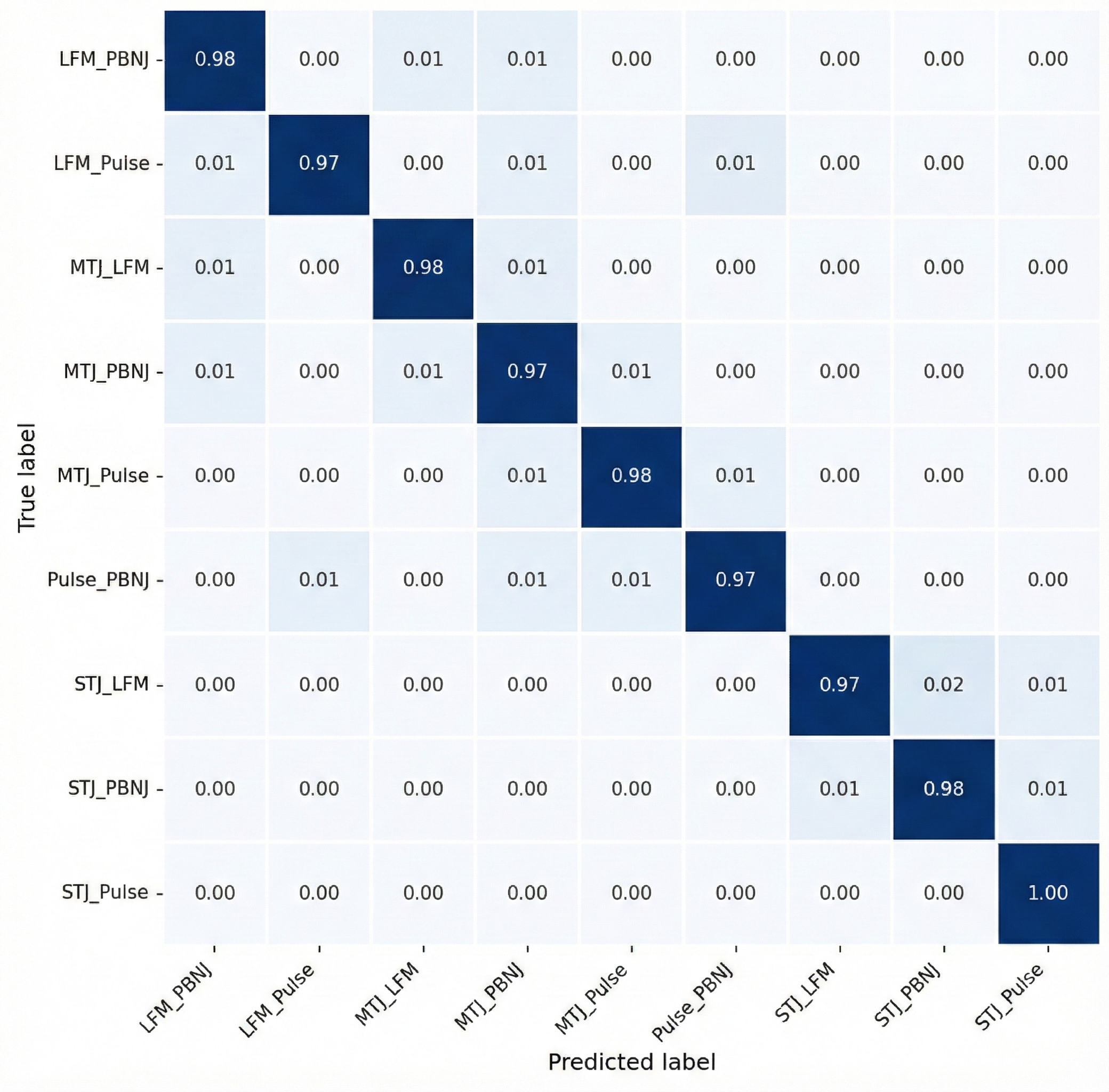}
    \caption{Confusion Matrix of the proposed SKANet.}
    \label{fig:cm}
\end{figure}

To further analyze the capability of the model to disentangle specific compound interference types, Fig. \ref{fig:cm} displays the confusion matrix of SKANet aggregated across the test set. The matrix exhibits high diagonal dominance and verifies excellent precision for compound signals.

Specifically, the STJ\_Pulse class is identified with near-perfect accuracy. This high discriminability is attributed to the distinct morphological differences between the two components where the sharp spectral lines of Single-Tone Jamming and the temporal bursts of Pulse Jamming are effectively captured by the frequency-domain and time-domain streams of our network respectively.

Minor misclassifications are observed primarily in classes involving wide-band noise such as the confusion between LFM\_PBNJ and MTJ\_PBNJ. In these scenarios, the high-power Partial-Band Noise Jamming component tends to mask the underlying sweep or tone structures of the secondary signal and creates spectral ambiguity. However, even in these challenging cases, the confusion rate remains below 5\% which demonstrates that the adaptive modality fusion mechanism effectively mitigates the masking effect by recalibrating the importance of Time-Frequency Images and Power Spectral Density features.

Table \ref{tab:metrics} provides a detailed quantitative breakdown of Precision, Recall, and F1 Scores across all evaluated methods. The results are averaged over the operational Jamming-to-Noise Ratio range of $-20$ dB to $15$ dB. SKANet achieves the highest average F1 score of 96.48\% and outperforms the runner-up method TFPENet by approximately 0.9\%.

This performance advantage is consistent across challenging compound classes involving diverse signal scales. For the MTJ\_Pulse category which combines discrete frequency tones with transient time-domain bursts, SKANet achieves a superior F1 score of 97.76\% and surpasses TFPENet by 0.81\%. Similarly, SKANet leads with an F1 score of 96.46\% for the STJ\_PBNJ category. These class-specific improvements validate the core design philosophy of the SK-ACB module where the network dynamically adjusts the receptive field to simultaneously attend to the micro-scale features of pulse signals and the macro-scale spectral characteristics of continuous wave interference. This capability ensures robust classification even in heterogeneous signal mixtures.

\subsection{Ablation Study}
\label{subsec:ablation}

To rigorously verify the contribution of each component within the proposed SKANet, we conducted a comprehensive ablation study. We constructed three reduced models by systematically removing key modules:
\begin{enumerate}
    \item w/o SK-ACB: The Multi-Branch SK-ACB module is replaced by standard convolutional blocks with static receptive fields, aimed at verifying the efficacy of the adaptive receptive field mechanism.
    \item w/o PSD Stream: The PSD branch is removed, and the network operates as a single-stream architecture using only TFI inputs, aimed at validating the necessity of dual-domain fusion.
    \item w/o SE Fusion: The SE attention module is removed, and features from the two streams are fused via simple concatenation, aimed at assessing the importance of adaptive channel recalibration.
\end{enumerate}

The experimental results are summarized in Table \ref{tab:ablation}.

\begin{table}[!t]
\renewcommand{\arraystretch}{1.3}
\caption{Ablation Study of SKANet Components (Averaged over JNR $-20$ dB to $15$ dB)}
\label{tab:ablation}
\centering
\resizebox{0.95\columnwidth}{!}{
\begin{tabular}{l c c c c c}
\toprule
\multirow{2}{*}{Models} & \multicolumn{4}{c}{Components} & \multirow{2}{*}{OA (\%)} \\
\cmidrule(lr){2-5}
 & STFT & PSD & SK-ACB & SE-Fusion & \\
\midrule
w/o SK-ACB    & \checkmark & \checkmark & $\times$   & \checkmark & 93.45 \\
w/o PSD Stream& \checkmark & $\times$   & \checkmark & N/A        & 94.82 \\
w/o SE Fusion & \checkmark & \checkmark & \checkmark & $\times$   & 96.15 \\
\textbf{SKANet} & \checkmark & \checkmark & \checkmark & \checkmark & \textbf{96.99} \\
\bottomrule
\end{tabular}
}
\end{table}

As observed from Table \ref{tab:ablation}, the SKANet model achieves the highest performance with an OA of 96.99\%, consistently outperforming all reduced versions. The most significant performance degradation is observed in the w/o SK-ACB model, where the replacement of the SK-ACB module with standard convolutions leads to a substantial accuracy drop of 3.54\%. This empirical evidence confirms that static kernels fail to simultaneously capture the conflicting features of compound interference, such as micro-scale pulse bursts and macro-scale sweep trends, thereby validating the critical role of the adaptive receptive field mechanism.

Furthermore, the exclusion of the PSD stream (w/o PSD Stream) results in an accuracy decline to 94.82\%. This degradation highlights the necessity of the dual-stream architecture, particularly in scenarios where fine-grained time-frequency textures are masked by wide-band noise, making the statistical stability of PSD an essential complementary feature. Finally, the w/o SE Fusion model, which employs a simple concatenation strategy, yields an accuracy of 96.15\%, lagging behind the proposed method by 0.84\%. This demonstrates that the SE mechanism effectively recalibrates channel importance, suppressing less informative features during modality fusion to further refine the global representation.

\section{Conclusion}
\label{sec:conclusion}

In this paper, we have addressed the challenge of classifying compound GNSS interference in dynamic and contested electromagnetic environments. We have recognized the limitations of existing single-domain Deep Learning models in resolving the scale ambiguity of entangled signals and proposed a cognitive dual-stream framework named SKANet. Our approach constructs a feature space robust to noise masking by combining the temporal localization capabilities of Time-Frequency Images with the statistical stability of Power Spectral Density.

The primary contribution of SKANet lies in the Multi-Branch SK-ACB module which functions as an adaptive spectral filter. The network successfully disentangles heterogeneous jamming components ranging from transient pulses to continuous sweeps within a unified architecture by dynamically modulating receptive fields via a selective attention mechanism. Extensive empirical evaluations on a large-scale synthetic dataset have demonstrated the superiority of our method. SKANet achieves a state-of-the-art Overall Accuracy of 96.99\% across the effective Jamming-to-Noise Ratio range of $-20$ dB to $15$ dB and significantly outperforms mainstream Convolutional Neural Network and Transformer baselines. Notably, the proposed framework exhibits high robustness in low-SNR regimes and maintains high discriminability even when signal features are heavily corrupted by thermal noise.


\end{document}